\documentclass[12pt]{article}

\usepackage{graphicx}
\graphicspath{{./figs/}}
\usepackage{subcaption}

\usepackage{booktabs,caption,fixltx2e}
\usepackage[flushleft]{threeparttable}

\usepackage[bold,full]{complexity}
\usepackage{booktabs}
\usepackage{multirow}
\usepackage{latexsym}
\usepackage[english]{babel}

\usepackage{longtable}
\usepackage{colortbl}
\usepackage{slashbox}

\usepackage{adjustbox,lipsum}
\usepackage{float}
\usepackage{caption}

\usepackage{morefloats}

\usepackage[cmex10]{amsmath}
\usepackage{amsfonts}
\usepackage{amsthm}
\usepackage{amssymb}
\usepackage{mathtools}
\DeclareMathOperator*{\argmin}{arg\,min}

\newtheorem{definition}{Definition}

\newtheorem{example}{Example}
\newtheorem{remark}{Remark}

\usepackage{color}

\usepackage[lined,ruled,commentsnumbered]{algorithm2e}

\usepackage{array}

\makeatletter
\newcommand{\removelatexerror}{\let\@latex@error\@gobble}
\makeatother

\begin{document}
\vspace*{0.35in}

\begin{flushleft}
{\Large
\textbf\newline{Transfer Learning based Dynamic Multiobjective Optimization Algorithms}
}
\newline
\\
Min JIANG \textsuperscript{1},
Zhongqiang HUANG \textsuperscript{1},
Liming QIU \textsuperscript{1},
Wenzhen HUANG \textsuperscript{2},
Gary G. YEN \textsuperscript{3, *}

\bigskip
\bf{1} Department of Cognitive Science and Technology, Xiamen University, China
\\
\bf{2} Institute of Automation of the Chinese Academy of Sciences, China
\\
\bf{3} School of Electrical and Computer Engineering, Oklahoma State University, USA
\\
\bigskip
* Corresponding authors: gyen@okstate.edu

\end{flushleft}

\section*{Abstract}
One of the major distinguishing features of the Dynamic Multiobjective Optimization Problems (DMOPs) is that optimization objectives will change over time, thus tracking the varying Pareto-Optimal Front (POF) becomes a challenge. One of the promising solutions is reusing ``experiences'' to construct a prediction model via statistical machine learning approaches. However, most existing methods neglect the non-independent and identically distributed nature of data to construct the prediction model. In this paper, we propose an algorithmic framework, called Tr-DMOEA, which integrates transfer learning and population-based evolutionary algorithms (EAs) to solve the DMOPs. This approach exploits the transfer learning technique as a tool to generate an effective initial population pool via reusing past experience to speed up the evolutionary process, and at the same time any population based multiobjective algorithms can benefit from this integration without any extensive modifications. To verify this idea, we incorporate the proposed approach into the development of three well-known evolutionary algorithms, nondominated sorting genetic algorithm II (NSGA-II), multiojective particle swarm optimization (MOPSO), and the regularity model-based multiobjective estimation of distribution algorithm (RM-MEDA). We employ twelve benchmark functions to test these algorithms as well as compare them with some chosen state-of-the-art designs. The experimental results confirm the effectiveness of the proposed design for DMOPs.

\textbf{Keywords:} Dynamic multi-objective optimization, Domain adaption, Dimensionality reduction, Transfer learning, Evolutionary Algorithm.


\section{Introduction}
%
%

%
%
%
%

One of the essential characteristics of Dynamic Multiobjective Optimization Problems (DMOPs) \cite{Farina_2004} is that objective functions will vary over time or under different environments. This underlying problem characteristic bears significant implications for real-world applications \cite{Cruz_2010}. A good example is dynamic portfolio optimization problem, which is common in deregulated electricity markets in which the operations of different power stations are controlled and coordinated to maximize profit while minimizing risk. There are various uncertainties in a deregulated electricity market, including spot market prices, load obligations, and strip/option prices \cite{XuLuhWhiteEtAl2006}. The values for some of these factors change over time, and it is ordinary to optimize for the market price every hour. However, the optimization approaches including population-based metaheuristics often find extreme difficulty to address the challenge since that the POF of a DMOP may change when the environment changes. Solving the DMOPs efficiently and effectively has become an important research issue in evolutionary computation community \cite{Nguyen_2012,daneshyari2012cultural}.

In recent years, a great deal of progress has been made and different types of algorithms have been proposed. In all of these methods, one class of approaches, the prediction based, has gained much attention. This class of approaches allows evolutionary algorithm (EA) and machine learning to be seamlessly integrated. After deriving a prediction model via machine learning techniques, the EAs can sustain the needed performance even if the environment changes over time. For example, in \cite{Sim_es_2013}, the authors proposed a memory-based EA which introduced two kinds of prediction models. The first one used the linear/nonlinear regression model to predict when the environment would change while the second model was based on Markov chains which was used to forecast changes. In \cite{Rossi_2008}, the authors suggested integrating motion information into an EA, such that the algorithm can track a time-changing optimum. In \cite{Stroud_2001}, the authors proposed a Kalman-extended genetic algorithm, and this algorithm was developed to determine when to re-evaluate an existing individual, when to produce a new individual, and which individual to re-evaluate.

The basic idea of these methods is ``keeping track of good (partial) solutions in order to reuse them under periodically changing environment'' \cite{Rossi_2008}. If we consider this view from a statistical point of view, this idea implies that the solutions of a dynamic optimization problem obey an identical distribution. In other words, the solutions which are used to construct the prediction model and the solutions forecasted by the prediction model meet the Independent Identical Distribution (IID) hypothesis to some extent. This assumption undoubtedly simplifies the complexity of the problem, however we have to understand there is an appreciable difference between the good, but out-of-date solutions and the proper and newly generated solutions, especially under a dynamic environment. That is to say, the changing POF may lead to the different distributions of the training samples and the predicted samples, and this problem is very difficult for the traditional machine learning methods. 

The findings from machine leaning community \cite{Pan.2010} already showed that a prediction model built by traditional machine learning methods leaves much room to be desired when the training samples and the predicted samples fail to meet the IID hypothesis. Transfer learning \cite{Pan.2010} allows the distribution of data used in training and testing to be different and it is becoming a useful weapon to overcome this difficulty. Therefore, the dynamic multiobjective optimization algorithms based on traditional machine learning methods, especially the prediction based algorithms, can also have significant performance improvements by overcoming the limitation caused by the IID, and transfer learning approach is a powerful tool we can use to improve performance of EAs for DMOPs.

In this paper, we argue that integrating transfer leaning approaches \cite{Pan.2010} into an EA can offer significant benefits to performance and robustness for designing better Dynamic Multiobjective Evolutionary Algorithms (DMOEAs). We adopt a domain adaptation method\footnote{a branch of transfer learning.}, called transfer component analysis \cite{pan2011domain}, to construct a prediction model. This model uses the gained knowledge of finding Pareto optimal solutions, but not the population, to generate an initial population pool for the optimization function at the next time. Based on this initial population pool, the optima of the changed environment can be found more efficiently and effectively. The proposed domain adaptation learning approach can be easily incorporated into any evolutionary-based multiobjective optimization algorithms. Please note that in this research, the dynamic refers to that objective functions will vary over time or under different environments.

Indeed, how to detect and identify dynamic changes is a crucial part of solving dynamic multiobjective optimization problems. However, in this paper, our focus is placed solely on how EA can quickly re-optimize a given dynamic optimization problem once the change is been detected and identified. This is in a similar spirit as those studies in fault tolerant control where focus is placed exclusively on designing a controller capable of accommodating the dynamic changes, leaving the fault detection and identification to be addressed separately.

The contribution of this research is the integration between transfer learning and classical evolutionary multiobjective optimization algorithms. This combination provides two benefits. \textbf{First}, the advantages of the EAs are preserved in the improved design for DMOPs. \textbf{Secondly}, the proposed design can significantly improve the search efficiency via reusing past experience which is critical for solving the DMOPs. An algorithm requires too much computing resources, often making it difficult to solve large-scale problems. The experiments also validate the assumption that the population plays a very important role for tracking dynamic optima, and \cite{Dang_2015,dang2016populations} proves it from the theoretical point of view. 

The rest of this paper is organized as follows: In Section \ref{sec:Preliminaries-and-Related}, we will introduce some basic concepts of dynamic optimization problems first and then discuss the existing works in this field. At the beginning of Section \ref{sec:TL-DMOA}, we will present some background on transfer learning, domain adaptation learning, and then introduce the transfer component analysis method in detail. After that we will propose the Transfer learning based Dynamic Multi-Objective Evolutionary optimization Algorithm, Tr-DMOEA. In Section \ref{sec:experiments}, we will present the experimental results of incorporating our approach to improve three well-known MOEAs: NSGA-II, MOPSO and RM-MEDA, specifically for DMOPs and all of the algorithms were tested on the IEEE CEC 2015 benchmark problems set. In Section \ref{sec:conclusion}, we will draw a summary of this paper and outline the future research directions.

\section{Preliminary Studies and Related Research}
\label{sec:Preliminaries-and-Related}

\subsection{Dynamic Multi-objective Optimazation}

Formally, a dynamic multiobjective optimization problem is defined as:
$$\textbf{Minimize}~F\left(x,t\right)=\left<f_{1}\left(x,t\right),f_{2}\left(x,t\right),...,f_{M}\left(x,t\right)\right>$$
$$s.t.\ x\in\Omega$$
where $x=\left<x_{1},x_{2},\ \ldots,\ x_{n}\right>$ is the decision vector and $t$ is the time or environment variable. $f_{i}\left(x,t\right):\Omega\ \rightarrow\ \mathbb{R}\ \left(i=1,\ \ldots,\ M\right)$.  $\Omega = \left[L_1,U_1\right] \times \left[L_2,U_2\right] \times \cdots \times \left[L_n,U_n\right]$. $L_i, U_i \in \mathbb{R}$ are the lower and upper bounds of the $i$-th decision variable, respectively. Please note that dynamic environments can be classified in different ways. For an in-depth description, the readers are referred to \cite{Sim_es_2013}.

\begin{definition}{\emph{[Dynamic Decision Vector Domination]}}
	At time $t$ , a decision vector $x_{1}$ Pareto dominate another vector $x_{2}$, denoted by $x_{1}\succ_{t}x_{2}$, if and only if\emph{:}
	\begin{equation}
	\begin{cases}
	\forall i=1,\ldots,M, & f_{i}(x_{1},t)\leq f_{i}(x_{2},t)\\
	\exists i=1,\ldots,M, & f_{i}(x_{1},t)<f_{i}(x_{2},t)
	\end{cases}.
	\end{equation}
\end{definition}

\begin{definition}{\emph{[ Dynamic Pareto-optimal Set ]} }
	Both $x$ and $\ensuremath{x^{*}}$ are decision vectors, and if a decision vector $x^*$ is said to be nondominated at time $t$ if and only if there is no other decision vector $x$ such that $x \succ_t x^*$ at time $t$. The Dynamic Pareto-Optimal Set (DPOS) is the set of all Pareto optimal solutions at time $t$, that is\emph{:}
	$$DPOS=\left\{ x^{*}|\not \exists x,~x\succ_{t}x^{*}\right\}.$$
\end{definition}

\begin{definition}{\emph{[Dynamic Pareto-optimal Front]}}
	At time $t$, the Dynamic Pareto-Optimal Front (DPOF) is the corresponding objective vectors of the DPOS.	
	$$DPOF=\left\{ F\left(x^{*},t\right)|x^{*}\in DPOS\right\}.$$
\end{definition}

For an ideal dynamic multiobjective algorithm, it must be able to find a set of solutions as close as possible to the changing Pareto-optimal Front and at the same time, the set of solutions should be as diverse as possible.

\subsection{Related Works}
Much progress \cite{Nguyen_2012,Raquel_2013,Azzouz_2016,Goh} has been made in the DMOPs field in recent years, and most existing algorithms can be classified into the following categories: Increasing/Maintaining Diversity methods, Memory based methods, Multi-population based methods, and Prediction based methods.

The increasing diversity methods tend to add variety to the population by using a certain type of methodology when the environment change was detected. For example, Cobb \textit{et al.} proposed the triggered hypermutation method \cite{cobb1990investigation}, and the basic idea of this method is that when change is identified, the mutation rate would be increased immediately, and this would make the converged population divergent again. This approach calls for some improvements, and one of them is that the mutation rate is in a state of uncontrolled change during the whole process, and this ultimately results in reduced performance of the algorithm. Therefore, Vavak\textit{ et al.} \cite{Vavak_1996} presented a mutation operator, called variable local search (VLS), to address the problem. The strategy that the VLS adopted was to gradually increase the mutation rate. Yen \textit{et al.} \cite{Woldesenbet_2009} proposed a dynamic EA which relocates the individuals based on their change in function value due to the change in the environment and the average sensitivities of their decision variables to the corresponding change in the objective space. This approach can avoid the drawbacks of previous methods to a certain extent.

Most of the methods in the maintaining diversity category assume that avoiding population convergence can help the algorithm track the changing optimum as soon as possible, and maintain diversity as one of the effective means to that end. Grefenstette \cite{Grefenstette92geneticalgorithms} proposed a Random Immigrants Genetic Algorithm (RIGA), and the method replaces some individuals in the population randomly. The idea of the RIGA is that introducing new genetic materials into the population can avoid the whole population converging toward a small area in the process of evolution. However, the drawback of the primitive immigrant method was the fitness values of the introduced individuals were usually low, so large amounts are eliminated during the selection stage, and as a result, it is very difficult to introduce different genes into the population. For solving this problem, Yang \cite{Yang_2007, Mavrovouniotis_2013} proposed the hybrid immigrants scheme, memory-based immigrants \cite{Yang_2008} and elitism-based immigrants \cite{Yang_2008}, and these methods are effective for dealing with periodically changing DMOPs. However, if the knowledge about the dynamic environment is limited, they would obtain a greatly reduced efficiency.

Dynamic NSGA-II (DNSGA-II) \cite{Deb}  proposed by Deb \textit{et al.} also shares a similar idea, and this method handles the DMOPs by introducing diversity when change is detected. There are two versions of the proposed DNSGA-II and they are respectively known as DNSGA-II-A and DNSGA-II-B. In  the DNSGA-II-A, the population is replaced by some individuals with new randomly created solutions, while in the DNSGA-II-B, diversity was guarded by replacing a percentage of the population with mutated solutions.

Memory mechanism enables EAs to record past information, and when it detects changes have occurred, stored information can be reused to improve the performance of the algorithm. Existing research showed that memory-based approaches tend to be more effective on the DMOPs with periodically changing environments.

Branke \cite{Branke} proposed a direct memory scheme where the best individuals in the population will be saved in an archive, and when the algorithm detects a change, those saved individuals can be retrieved and returned to the population to replace the same number of individuals. In \cite{Chi_Keong_Goh_2009}, the author proposed a co-evolutionary multiobjective algorithm which hybridizes competitive and cooperative mechanisms to solve the DMOPs. In this algorithm, the out-of-date archived solutions are replaced by an external population. In \cite{Wang_2009}, the authors presented an algorithm called MS-MOEA to tackle the challenges of DMOPs. In the method, adaptive genetic and differential operators were used to speed up the convergence speed and a Gaussian local search operator was employed to prevent from premature convergence. At the same time the fast hyper-volume strategy \cite{while2006faster} was proposed to achieve a better starting population when changes occur frequently. The above methods meet some problems, e.g. slow convergence and poor diversity, when the environment changes. As a result the authors in \cite{Azzouz_2015} proposed an adaptive hybrid population management strategy using memory, local search and random strategies to effectively handle environment dynamicity in DMOPs. The special feature of this algorithm is that it can adjust the number of memory and random solutions to be used according to the change severity.

The Multi-population strategy is considered as one efficient solution for the DMOPs, especially for the multiple peaks and the competing peaks problems. Branke \textit{et al.} \cite{Branke_2000} proposed the self-organizing scouts method, and this method splits the population into scout and base populations, and the two populations are responsible for exploitation and exploration respectively. In other words, the base population searches for the optimal solution and if the base population finds a peak, then the scout population is generated to track the change of this new peak. Li and Yang \cite{Li_2008} employed a multi-population particle swarm optimization (PSO) algorithm to solve multiple peaks problems. In their method, a population uses evolutionary programming, which shows a better global search ability when compared to other EAs, to explore the most hopeful areas in the whole search space, and at the same time, several subpopulations use the fast PSO algorithm to find the local optima. Yang \cite{Yang_2010} used hierarchical clustering technique to divide the population into different subpopulations, and the main advantage of this design is that the initial individuals of the subpopulations can be generated automatically according to the fitness landscape.

In general, a good dynamic optimization algorithm should be able to track the changing optimal solution even under high severity and frequency of changes. It must be able to reuse as much information available from previous generations to speedup the optimization search. As a result, in recent years the prediction-based DMOPs algorithms have received much attention. This class of methods predicts the state of the changing environment typically using  the information that already exists and some forms of machine learning techniques, and then makes a decision such that the algorithms can accommodate the changes in advance. This is one of the reasons why the prediction-based approaches can improve performance of an algorithm handling the DMOPs, compared with other types of approaches.

Bosman \cite{Bosman_2007} believed that the decision made at one point would affect the optima obtained in the future, so for the dynamic optimization problems, he proposed an algorithmical framework which integrated machine learning, statistic learning, and evolutionary computation, and this framework can effectively predict what the state of environment is going to be. In \cite{Rossi_2008}, the authors suggested that the state of an optimum should contain the location and the speed information, so the Kalman filter technique can be used to estimate the state of the system and the error. The authors proposed an EA to measure the state of the past optimum and then use the Kalman filter to obtain an estimated value of the optimum in the next time instance.

Stroud \cite{Stroud_2001} proposed the Kalman-extended Genetic Algorithm (KGA), and the basic idea of the KGA was that two types of uncertainties surrounded the estimated value of an individual in a dynamical environment. The first type of uncertainty is produced by the dynamic of the environment while the second type was related to the evaluation of individuals. For the different situations, the KGA has two different ways to update the covariances, and uses the Kalman filter technique to predict the two uncertainties which allows the algorithm to work well in a dynamic environment.

In \cite{Aimin_Zhou_2014}, Zhou \textit{et al.} presented an algorithm, called Population Prediction Strategy (PPS), to predict a whole population instead of predicting some isolated points. There are two key concepts here: center point and manifold. Whenever a change is detected, the algorithm uses a sequence of center points obtained from the search progress to predict the next center point, and at the same time, the previous manifolds are used to estimate the next manifold. The main problem of this method is that, it is difficult to obtain historical information at the beginning stage, and this may lead to poor convergence.

Recently, there are some works exploiting knowledge reuse techniques or machine learning in evolutionary computation that have been proposed. In \cite{iqbal2017cross}, the authors propose an approach based on transfer learning and genetic programming to solve complex image classification problems. The basic idea of the proposed algorithm is that the knowledge learned from a simpler subtask is used to solve a more complex subtask, and reusing knowledge blocks are discovered from similar as well as different image classification tasks during the evolutionary process. In \cite{iqbal2014reusing}, the authors present a genetic programming-like representation to identify building blocks of knowledge in a learning classifier system, and the proposed method can extract useful building blocks from simpler and smaller problems and reuse them to learn more complex multiplexer problem. In \cite{feng2015memetic}, the authors present an evolutionary memetic computing paradigm that is capable of learning and evolving knowledge meme that traverses two different but related problem domains, capacitated vehicle routing problem and capacitated arc routing problem, for greater search efficiency. Experimental results show that evolutionary optimization can benefit from this approach.”

However, we are not exactly sure if the data we using to construct the prediction model and the data we are going to predict by the above model obey a similar distribution. Conversely, the real-world applications repeatedly reminded us that, it is not wise to assume the IID hypothesis as a prerequisite, especially for the DMOPs. Unfortunately, most of the existing methods often assume that the solutions at different times have an IID structure, and we believe that this assumption is one of main reasons for the failure of existing DMOEA algorithms. After all, a poor prediction model is very likely to bring the search process to a hopeless place, which means actual results will be worse than a method which does not use predictive technique, if the prediction model turn out to be inaccurate. 

We believe that historical information about the POF or POS is very useful, and the reason is that for a DMOP, the POSs or POFs at different times may not be exactly the same, but they must be correlated. Therefore we conjecture that an appropriate use of the information extracted from the obtained POS or POF will bring great benefits to track the changing POF, but at the same time, we must admit the rationality and the generality of the assumption of non-independently and identically distributed data. From these basic points of view, we propose a framework which integrates transfer learning and EAs for solving the DMOPs. Two of the major advantages of the proposed approach are as follows: at first, the proposed method does not assume IID hypothesis as a prerequisite, and it is enabled to escape serious consequences of an unsuitable model. Secondly, this approach is designed to generate a population-building prediction model, so that any population-based optimization algorithms may benefit from this integration without any extensive modification.

\section{Transfer Learning based Dynamic Multi-objective Optimization Algorithm }
\label{sec:TL-DMOA}
In this section, we propose a transfer learning based dynamic optimization algorithm. Our motivation is that the solutions of a dynamic optimization problem under different environments obey  different  probability distributions, and these distributions are not identical but are correlated. If we can map these different distributions into a latent space, and in this space the distributions are as ``similar'' as possible, then we can use the available solutions to generate an initial population, such that the solutions under a new environment can be computed with low computational cost. In principle, this design  is a reuse process of the knowledge we already obtained.

Before giving the details of the proposed approach, we need to introduce background information of the domain adaptation learning we will use it in our design.

\subsection{Transfer Component Analysis}
Briefly speaking, Domain Adaptation Learning (DAL) \cite{BenDavid.2010, Patel.2014, jiang2017integration}, a branch of transfer learning, is to reuse the knowledge acquired from a source domain to perform a task in a target domain, which is related to, but distinct from the source domain. In the context of this research, a domain includes a sample space $\mathcal{X}$ and the corresponding marginal distribution $P(X)$, where $X=\{x_{1},x_{2},\ldots,x_{n}\} \subseteq \mathcal{X}$. We say that two domains are different, which means they have different sample spaces and the marginal distributions are different.

The researchers \cite{BenDavid.2007} believe that it is a promising solution using the DAL  to find a “good” representation to decrease the difference between the distributions of source and target domains. Gretton {\textit{et al.}} \cite{Gretton.2006} noted that the distance between two different distributions can be evaluated by a particular function, and in the Reproducing Kernel Hilbert Space (RKHS), the computational cost of the evaluation can be reduced. Based on this observation, Gretton {\textit{et al.}} \cite{Gretton.2006, Smola.2007} proposed a nonparametric distance estimation method called Maximum Mean Discrepancy (MMD) to differentiate distributions in the RKHS \cite{Steinwart.2002}. The MMD measures the discrepancy between two distributions by computing the difference of the mean values for the source domain and target domain. The advantages of the MMD approach is its simplicity and accuracy.

\begin{definition} (Maximum Mean Discrepancy \emph{\cite{Gretton.2006}}) \emph{:} Let $p$ and $q$ be two Borel probability measures defined on a domain $\mathcal{X}$; and $X= \{x_1,\cdots, x_m\}$ and $Y=\{y_1,\cdots, y_n\}$ be two observations drawn from $p$ and $q$ respectively. Let $\mathcal{F}$ be a class of functions $f: \mathcal{X} \rightarrow \mathbb{R}$, then the maximum mean discrepancy ($\mathbf{MMD}$) can be defined as\emph{:}
	$$\mathbf{MMD}(\mathcal{F},p,q)\coloneqq\underset{f\in\mathcal{F}}{\sup}\left(\frac{1}{m}{\displaystyle \sum_{i=1}^{m}f(x_{i})}-\frac{1}{n}\sum_{i=1}^{n}f(y_{i})\right).$$
\end{definition}

In a Reproducing Kernel Hilbert Space, $f$ can be written as $f(x)=\left<\phi(x),f\right>$, where $\phi(x):\mathcal{X}\rightarrow\mathcal{H}$. So the empirical estimate of $\mathbf{MMD}$ can be rewritten as:
\begin{equation}
\mathbf{MMD}(\mathcal{F},p,q)\coloneqq \left\|\frac{1}{m}{\displaystyle \sum_{i=1}^{m}\phi(x_{i})}-\frac{1}{n}\sum_{i=1}^{n}\phi(y_{i})\right\|^{2}_{\mathcal{H}}.
\label{MMD-2}
\end{equation}

By using the so-called ``kernel trick''\cite{shawe2004kernel},  we can rewrite Equation (\ref{MMD-2}) as

\begin{align}
\label{MMD-3}
\mathbf{MMD}(\mathcal{F},p,q) &\coloneqq \sum_{i=1}^{m}\sum_{j=1}^{n}\mathbf{tr}[\hat{K}(\frac{1}{m \times m}L_{ii}-\frac{1}{m\times n}L_{ij}\\ \nonumber
& \quad -\frac{1}{n \times m}L_{ji}+\frac{1}{n \times n}L_{jj})]\\ \nonumber
&\coloneqq \mathbf{tr}(\hat{K}L), \nonumber
\end{align}
where $\mathbf{tr}(A)$ refers to the trace of the matrix $A$, and the matrix
\begin{equation}
\hat{K}= \begin{pmatrix}
\hat{K}_{X,X} & \hat{K}_{X,Y} \\
\hat{K}_{Y,X} & \hat{K}_{Y,Y}
\end{pmatrix} \in \mathbb{R}^{(m+n)\times(m+n)}.\\
\label{hatkernelmatrix}
\end{equation}
$\hat{K}_{X,Y}$ is a kernel matrix with  $k_{i,j} = \kappa(x_i,y_j) = \phi(x_i)^T\phi(y_j)$, where $\kappa(\cdot,\cdot)$ is a kernel function and $\phi(\cdot)$ is a feature mapping function. This matrix reflects data similarity in the domains $X$ and $Y$. $\hat{K}_{X,X}, \hat{K}_{Y,X}$ and $\hat{K}_{Y,Y}$ have the similar meanings. Matrix $L$ contains the coefficients to scale matrix according to Equation (\ref{MMD-2}) and its elements are as follows.
\begin{equation}
L(i,j)=
\begin{cases}
& \frac{1}{m \times m}, \quad x_i, x_j \in X \\
& \frac{1}{n \times n}, \quad x_i,x_j \in Y\\
& -\frac{1}{m \times n}, \quad  otherwise
\end{cases}.
\label{matrixL}
\end{equation}

On the basis of the MMD, Pan \textit{et al}. proposed  a dimension reduction method \cite{pan2011domain} called Maximum Mean Discrepancy Embedding (MMDE) to (1) find a low-dimensional space to reduce the difference between source and target’s distributions as well as (2) to preserve the main statistical properties, maximization of data variance in the first extracted orthogonal components of the original data $X$ and $Y$. In MMDE, the kernel function $\kappa$ is learned (or optimized) from the data, which makes it computationally expensive, so the authors in \cite{pan2011domain} proposed other dimension reduction-based methods called Transfer Component Analysis (TCA) and its Semi-Supervised version of TCA, SSTCA, to transform the problem of learning an entire kernel matrix to a low-rank matrix $W$ instead.

Now let us consider how to obtain the martix $W$ by using the TCA method. Suppose that $W$ is a $(m+n) \times d$   matrix. For any vector $x$,  let $\phi(x) = W^T\kappa_x \in \mathbb{R}^d$, where $\phi(\cdot)$ is a feature mapping function. Let $\kappa_x = [\kappa(x_1, x),  \ldots, \kappa(x_m, x), \kappa(y_1, x), \ldots, \kappa(y_n, x)]^T$, and the matrix $\hat{K}$  in Equation (\ref{hatkernelmatrix}) can be transformed as follows.
\begin{align}
\hat{K} \nonumber &= [\phi(x_1), \ldots, \phi(x_m),\phi(y_1),\ldots,\phi(y_n)]^T \times\\
\nonumber&\quad [\phi(x_1), \ldots, \phi(x_m),\phi(y_1),\ldots,\phi(y_n)]       \\ \nonumber
&= [W^T\kappa_{x_1},\ldots,W^T\kappa_{x_m},W^T\kappa_{y_1},\ldots,W^T\kappa_{y_n}]^T \times\\ \nonumber
&\quad  [W^T\kappa_{x_1},\ldots,W^T\kappa_{x_m},W^T\kappa_{y_1},\ldots,W^T\kappa_{y_n}] \nonumber \\
&= [\kappa_{x_1},\ldots,\kappa_{x_m},\kappa_{y_1},\ldots,\kappa_{y_n}]^TWW^T \nonumber\\
&\quad [\kappa_{x_1},\ldots,\kappa_{x_m},\kappa_{y_1},\ldots,\kappa_{y_n}] \nonumber \\
& =  K^TWW^TK \nonumber \\
& = KWW^TK.  \qquad
\end{align}

Please note that the matrix $K$ is a symmetric matrix, so $K^T=K$, and then
$\mathbf{tr}(\hat{K}L) = \mathbf{tr}(KWW^TKL)$. According to the property of the trace of a matrix, we can rewrite Equation (\ref{MMD-3}) as follows.

\begin{align}
\mathbf{MMD}(\mathcal{F},p,q) = & \mathbf{tr}(\hat{K}L) \nonumber \\
= & \mathbf{tr}(KWW^TKL) \nonumber\\
= & \mathbf{tr}(W^TKLKW).
\end{align}

Now the optimization problem for the TCA algorithm can be written as follows:
\begin{equation}
\begin{aligned}
&\argmin_W
& &\mu \cdot \mathrm{tr}(W^{T}W)+\mathrm{tr}(W^{T}KLKW) \\
& \text{subject to}
& & W^{T}KHKW=\mathbf{I}\label{eq:final-obj},  \\
\end{aligned}
\end{equation}
{where $H=\mathbf{I}-\frac{1}{m+n}\mathbf{1}\mathbf{1}^{T}$ and $\mathbf{I}$ is a \,$(m+n)\times(m+n)$ identity matrix. $W^TW$ is a regularization term. $\mathbf{1}$ is a $(m+n)\times 1$ all-ones matrix.} $m$ and $n$ represent the numbers of samples in the source and target domains, respectively. $\mu$ is the tradeoff parameter. This optimization problem can be transformed into a trace maximization problem. According to the method presented in \cite{Scholkopft.1999}, the trace maximization problem can be solved by the Generalized Eigenvalue Decomposition (GED), and the solution is composed of the $d$ leading eigenvectors. The pseudo-code of TCA is given in Algorithm \ref{alg:TCA}. 

\begin{figure}[htbp]
	\centering
	\removelatexerror
	\begin{algorithm}[H]
		\label{alg:TCA}
		\caption{TCA}
		\LinesNumbered
		\KwIn{ \small{ Source domain $X$; target domain $Y$; a kernel function $\kappa(\cdot,\cdot)$;}}
		\KwOut{\small{ Matrix $W$}}
		
		Construct the Kernel Matrix $\hat{K}$, the Matrix $L$, and the Matrix $H$ according to (\ref{hatkernelmatrix}), (\ref{matrixL}) and (\ref{eq:final-obj}) \;
		
		Construct the Matrix $W$ by using the $d$ leading eigenvectors of $(KLK+\mu I)^{-1}KHK$ \;		
		
		\Return the matrix $W$\;
	\end{algorithm}
\end{figure}

\subsection{Tr-DMOEA}
Dynamic multiobjective optimization problem is a computationally expensive task. This implies that it requires a lot of computational resources to search for the varying POS at a certain time. If the knowledge about the POS and POF can be reused to predict future POFs or POSs under different environments, this usually implies performance improvement as well as less computational resource consumption. As a result, we believe that the prediction-based dynamic multiobjective optimization algorithm presents a promising solution.

However, the existing algorithms generally neglect the assumption of Non-Independent Identically Distributed (Non-IID), and it is obvious that the individuals under different environments obey different distributions. This also means that those dynamic optimization algorithms based on the traditional machine learning approach leave much room for improvement. So we put forward the use of the domain adaptation technique to develop a novel DMOEA. 

The approach developed is to map different distributions that the solutions obey at different times into a new latent space via the domain adaptation method. In the latent space, the MMD value of different distributions will be as small as possible while variance of the data will be kept the same. In other words, we will make those distributions that the solutions under different environments obey  as similar as possible in the latent space, so we can map the POF we have obtained into the space, and then use those mapped solutions to construct a population which will be used to search for the POF under a new environment.

In the following Tr-DMOEA algorithm, $F_t$ is the current dynamic optimization function assuming its POF has already been found. $F_{t+1}$ is the optimization function at the next time. The major part of the algorithm, Tr-IPG, utilizes the POF at time $t$ and the transfer learning method to generate a population which can be used to search for the POF at time $t+1$. More specifically, we take the obtained Pareto-optimal Front (POF) at time $t$ as a source domain; the feasible solutions of the next time, time $t+1$, as the target domain, and then construct a mapping function $\varphi$ by using the domain adaptation approach. This mapping function will embed the distributions that the source and target domain obey separately into a latent space, and in that space the difference between the two distributions will become as small as possible. From this, we can use the POF already found to generate an initial population which can be used to search for the POF of the next moment.

\begin{figure}[!htb]
	\centering
	\removelatexerror
	\begin{algorithm}[H]
		\label{alg:trdmoo}
		\caption{Tr-IPG: Transfer Learning based Initial Population Generator}
		\LinesNumbered
		\KwIn{ \small{ The Dynamic Optimization Function $F_{t+1}(\cdot)$; the POF of the function $F_{t}(\cdot)$ at time $t$, $POF_{t}$ = $\{p_1, \ldots p_m\}$; a kernel function $\kappa(\cdot,\cdot)$.}}
		\KwOut{\small{ A population \bf{Pop-init}}.}
		
		Initialization\;		
		
		For the optimization functions $F_t(\cdot)$  and $F_{t+1}(\cdot)$, randomly generate two sets of the solutions $X_s $ and $Y_t $ \tcc*{Remark \ref{tr-ipg-remark-1}}
		
		Calculate the objective values of the optimization functions $F_{t}(X_s)$ and $F_{t+1} (Y_t)$;
		
		W $\leftarrow$ TCA($\{F_{t}(X_s)\}$, $\{F_{t+1} (Y_t)\}$, $\kappa$)\;
		
		$PLS$ $\leftarrow \emptyset$; \tcc{Remark \ref{tr-ipg-remark-2}}
		
		\For{ every $p$ $\in$ $POF_{t}$ }{ \label{alg: start}
			$\kappa_p \leftarrow \left[\kappa\left(F_t(X_s(1)), p\right), \ldots, \kappa\left(F_{t+1}(Y_t(n_t)), p\right)\right]^T$
			
			$\varphi(p) \leftarrow W^T\kappa_p$\;
			$PLS$ = $PLS \cup \{ \varphi(p)$\}\;
			
		}
		
		\For {every $l$ $\in PLS$} {
			
			$x \leftarrow \textrm{arg}\underset{x}{\min}\lVert  \varphi \left( F_{t+1}\left(x \right) \right) - l \rVert$\tcc{Remark\ref{tr-ipg-remark-3}}	
			
			\bf{Pop-init} = \bf{Pop-init} $\cup \lbrace x \rbrace$ \ ;
		}

		\Return \bf{Pop-init} \label{alg:end}\;
	\end{algorithm}
\end{figure}

\begin{figure*}[!htb]
	\centering
	\includegraphics[width=0.95\textwidth]{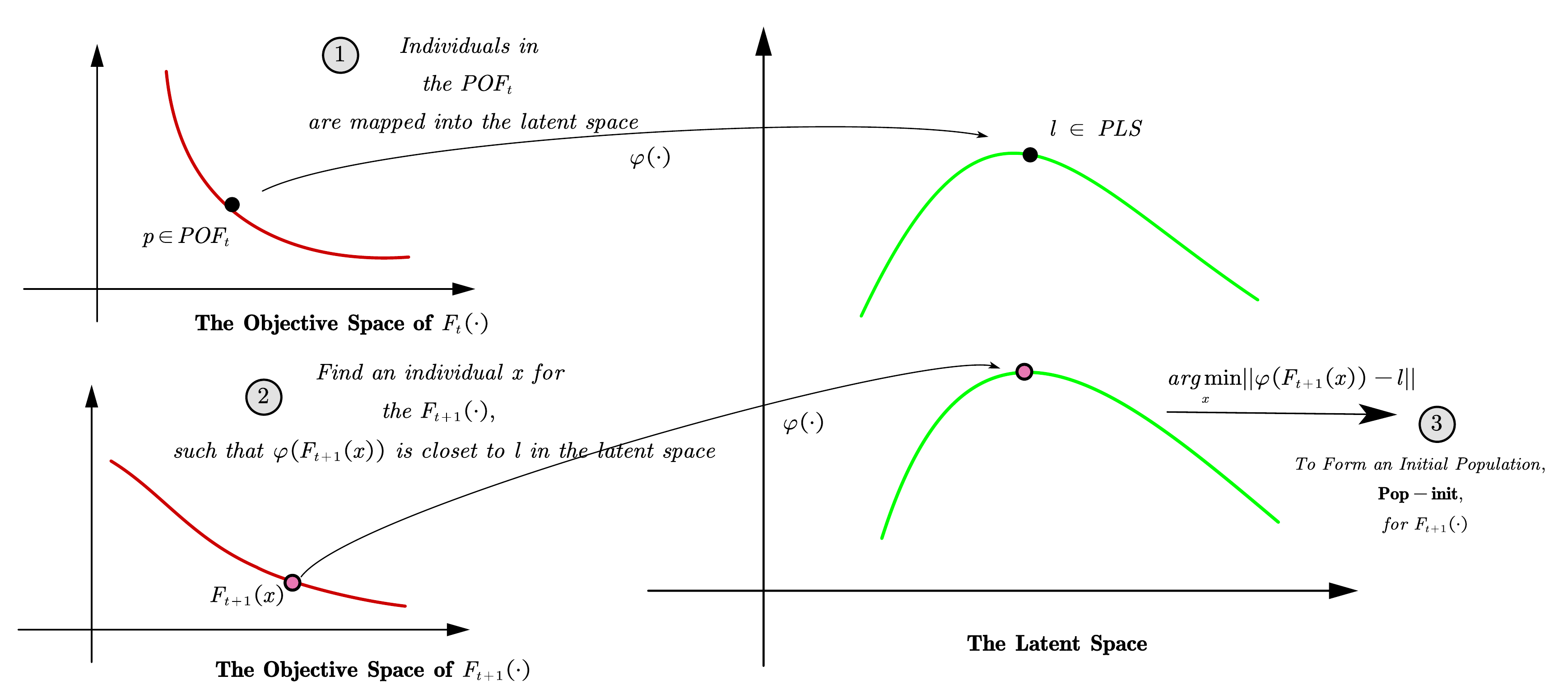}
	\caption{The key steps of the Tr-IPG algorithm. Step 1: Map the obtained POF into the latent space; Step 2: Find individuals for $F_{t+1}(\cdot)$; Step 3: Generate initial population pool for the problem $F_{t+1}(\cdot)$.}\label{fig:digit}
	\label{Pic:TR-IPG}
\end{figure*}

In order to help the readers quickly grasp the basic idea of the algorithm Tr-IPG,  Figure \ref{Pic:TR-IPG} has been presented to illustrate the key elements of the algorithm. This diagram describes the operational process from Line \ref{alg: start} to Line \ref{alg:end} of the Tr-IPG.

Please note that the input to the TCA are the samples from the solutions at time $t$ and $t+1$, and its output is a transformation matrix $W$. We can use the matrix $W$ to construct the latent space. The Step 1 (i.e., the upper left corner of the diagram) depicts the process of mapping the POF at time $t$ into the latent space, and the Steps 2 and 3 describe how to search for an initial population which can be used to solve the dynamic optimization at time $t+1$.

\begin{remark}
	\label{tr-ipg-remark-1}
	The numbers of the elements of $X_s$ and $Y_t$ are predefined. Let $\left|X_s\right| =n_s$ and $\left|Y_t\right| =n_t$. In general, more sampling often means  a better result, but it also needs to pay a higher computational cost, so the decision about how many solutions needed to be produced in this step depends on the resources available.
\end{remark}

\begin{remark}
	\label{tr-ipg-remark-2}
	$PLS$ is the acronym of Particle in the Latent Space and it can be regared as a set of the mapped solutions in the latent space. 
\end{remark}

\begin{remark}
	\label{tr-ipg-remark-3}
	We want to find a decision variable $x$, such that in the latent space, $\varphi(F_{t+1}(x))$ is closet to $l \in PLS$ in the latent space. This also means that we need to solve a single objective optimization problem here, and any single objective  optimization algorithm can be applied to solve the problem. In this research, we use the Interior Point Algorithm to solve the problem.
\end{remark}

What the Tr-IPG algorithm outputs is a population, so it is not difficult to find that we can combine any type of population-based optimization algorithms with the Tr-IPG to obtain a transfer learning based dynamic multiobjective EA.

\begin{remark}
	For the TCA algorithm, the major time is spent on eigenvalue decomposition. It takes $O(d(m_1+m_2)^2)$ time when $d$ nonzero eigenvectors are to be extracted, where $m_1$ and $m_2$ are the numbers of the solutions which are generated to construct the latent space. The Tr-IPG spends $O(n_1)$ time to map an individuals in the $POF_t$ to the latent space and we use the Interior Point Algorithm to find an indidvidual in the objective space of $F_{t+1}(\cdot)$. For the primal dual interior point method, suppose the constraint matrix $A$ has $n$ rows and $m$ columns, and $n < m$, it has $O(\sqrt{m}L)$ iterations and $O(m^3L)$ arithmetic operations, where $L$ is total number of bits of the input.
\end{remark}

\begin{example}
	A specific numerical example is helpful in understanding how the Tr-IPG algorithm works. For example, we employ the Tr-IPG algorithm to solve the FDA4 problem, which is a three-objective dynamic optimization problem. Let us suppose that the POF of the FDA4 problem \cite{helbig2015benchmark} at time $t$, $POF_t$, has been found and $p \in POF_t$. The Tr-IPG uses the TCA algorithm to obtain a mapping function $\varphi(\cdot)$, and this mapping function is utilized to map the $p$ into a twenty-dimensional latent space\footnote{The dimensionality of the latent space depends on the parameters of the TCA algorithm.}, and it means that $l=\varphi(p)$ is a twenty-dimensional vector. After that the Tr-IPG algorithm will find a solution $x$ for the FDA4 problem at time $t+1$, and this solution $x$ satisfies the requirement that it is nearest to $l$ in the latent space. The Tr-IPG will output the solution $x$ as one of the individuals of the initial population which can be used to solve the FDA4 problem at time $t+1$.
\end{example}

\begin{figure}[htbp]
	\centering
	\removelatexerror
	\begin{algorithm}[H]
		\label{alg:trdmoo-imple}
		\caption{Tr-DMOEA: Transfer Learning based Dynamic Multi-objective Evolutionary Algorithm }
		\LinesNumbered
		\KwIn{ \small{ The Dynamic Optimization Function $F(X)$; a population based multiobjective algorithm MOA; a kernel function $\kappa(\cdot,\cdot)$}.}
		\KwOut{\small{ the POFs of $F(X)$}.}
		
		Initialization \;
		
		Use {\bf{MOA}} to solve $F_0(X)$ to get a $POF_0$ \;
		
		\For {$t = 1$ \KwTo $n$} {
			
			Next-Pop = Tr-IPG$( F_t(\cdot), POF_{(t-1)}, \kappa(\cdot,\cdot) )$ \tcc*{When a change occurred, we use Tr-IPG to generate an init population.}
			
			$POF_t$ = {\bf{MOA}}(Next-Pop) \;
			
			\Return $POF_t$ \;
			
		}
		
	\end{algorithm}
\end{figure}

\section{Empirical Study}
\label{sec:experiments}
Practically speaking, the proposed approach is compatible with any type of population-based optimization algorithms. As a case study, in our experiments, we select three well-represented algorithms with different operating metaphors to verify our approach. The first one is the NSGA-II \cite{deb2002fast} and it is a multiobjective genetic algorithm that applies nondominated sorting and crowding distance. The second multiobjective optimization algorithm is based on particle swarm optimization and it is simply called as MOPSO \cite{coello2002mopso}. The third one is the RM-MEDA \cite{zhang2008rm}, which is a regularity model based multiobjective estimation of distribution algorithm.

The three corresponding algorithms with the proposed transfer learning  are called Tr-NSGA-II, Tr-MOPSO, and Tr-RM-MEDA, respectively for dynamic optimization. It should be noted that the original designs, NSGA-II, MOPSO, and RM-MEDA are not appropriate for dynamic optimization. It is not difficult to find that these three algorithms belong to different categories, but all of them are well-developed, so it can strengthen the persuasive power and the confidence level to incorporate the proposed technology. At the same time, we also compare the new algorithms with other state-of-art designs.

One thing we need to emphasize is that in all our experiments, the parameters are set the same. In other words, for these twelve test functions and three different algorithms, we have used the same parameters and do not tune the parameters in TCA under different configurations for a better performance.

\subsection{Performance Metrics, Testing Functions and Settings}
In this research, we use four performance metrics, the Inverted Generational Distance (IGD) and its variants, the Reactivity measure (React) and its variants, to evaluate  the quality  of the solutions  obtained by these competing algorithms.

\begin{enumerate}
	\item The inverted generational distance (IGD) \cite{sierra2005improving} is a metric  to quantify the performance of a multiobjective optimization algorithm. Let $P^*$ be the set of uniformly distributed Pareto optimal solutions in the POF and $P$ represent the POF obtained by the algorithm,  the definition of the IGD is：
	\begin{equation}
	\label{def:igd}
	\mathrm{IGD}(P^*, P, C) = \frac{\sum_{v^* \in P^*}\min_{v\in P}\left\|v^*-v\right\|}{\left|P^*\right|}.
	\end{equation}
	
	If we want the value of IGD to be as small as possible, the $P$ should be close enough to $P^*$. In other words, the IGD depicts the difference between the ideal POF and the POF obtained by the competing algorithms.
	
	Please note that the definition of the IGD is slightly different from the original one, and the major difference is the parameter $C$ in Equation (\ref{def:igd}). The parameter $C$ is a combination of the benchmark functions parameters. We call it as configuration of the benchmark functions. The configurations we used in our experiments are described in Table \ref{def: different configs}. 
	
	\item One variant of the IGD, called MIGD, can also be used to evaluate dynamic multiobjective optimization algorithms \cite{Muruganantham_2015,Muruganantham_2016} , and it takes the average of the IGD values in some time steps over a run as the performance metric, given by
	\begin{equation} 
	\mathrm{MIGD}(P^*, P, C) = \frac{1}{|T|}\sum_{t \in T}\mathrm{IGD}(P_t^{*}, P_{t}, C),
	\end{equation}
	where $P_t^{*}$ and $P^{t}$ represent the points set of the ideal POF and the approximate POF obtained by the algorithm at time $t$. We also want to evaluate those algorithms under different environments, so a novel metric, DMIGD, is defined based on the MIGD, and the definition of the DMIGD is as follows:
	
	\begin{equation}
	\mathrm{DMIGD}(P^*, P, C) = \frac{1}{|E|}\sum_{C \in E}\mathrm{MIGD}(P_t^{*}, P_{t}, C),
	\end{equation}
	where $|E|$ is the number of the different environments experienced. In our experiments, we choose eight different configurations. As a result, $|E|$ equals to eight. What we want to point out is that the DMIGD can evaluate a dynamic optimization algorithm from a  high-level view and it bears a significant difference with the MIGD since the MIGD just considers the dynamics in one environment.
	
	
	
	\item The reactivity measure (React) \cite{sola2010parallel} is used to measure the robustness of an algorithm, and its definition is as follows:
	\small
	$$
	\mathrm{React}_\epsilon(t,C) = \min\left\{t'-t | t < t'\in\mathbb{N}, \frac{acc(t')}{acc(t)}\geq 1-\varepsilon\right\},
	$$
	\normalsize
	where $acc(t)=\frac{HV(POF(t))}{\max HV(POF)}$ implies the accuracy rate of computing the POF at time $t$, and $HV$ refers to the value of Hypervolume \cite{nebro2008abyss}. The React describes how quickly a dynamic optimization algorithm can recover from a change, or convergence speed after changes. The value of the React is the smaller the better. We also want to evaluate the algorithms on a macro-scale, so we derive two additional metrics based on the React,	
	$$
	\mathrm{MReact}_\epsilon(T,C) = \frac{1}{|T|}\sum_{t \in T} \mathrm{React}_\epsilon(t, C),
	$$
	\small	
	\begin{equation}
	\mathrm{DMReact}_\epsilon(T,C) = \frac{1}{|E|}\sum_{C \in E} \mathrm{MReact}_\epsilon(T, C).
	\end{equation}
	\normalsize	
\end{enumerate}

The MReact value can be considered as an average of the React values at different time points, but under the same configuration; DMReact is an average of the MReact values over different configurations considered.

In the experiments, we apply the IEEE CEC 2015 Benchmark problems set in Table \ref{Results:testfunctions} as test functions and the problem set has twelve testing functions \cite{helbig2015benchmark}. In the definitions, the decision variables are $x=(x_1,\ldots,x_n)$ and $t = \frac{1}{n_t}\left\lfloor \frac{\tau_T}{\tau_t} \right\rfloor$, where $n_t, \tau_T$, and $\tau_t$ are the severity of change, maximum number of iterations, and frequency of change, respectively. Table \ref{def: different configs} describes the different combinations of $n_t$, $\tau_t$, and $\tau_T$. Please note that for each $n_t$-$\tau_T$ combination, there will be $\frac{\tau_T}{\tau_t}$ environment changes. In other words, in all of our experiments, there are altogether twenty changes for the twelve dynamic problems.

\begin{figure*}[htbp]
	\centering
	\includegraphics[width=1.05\textwidth]{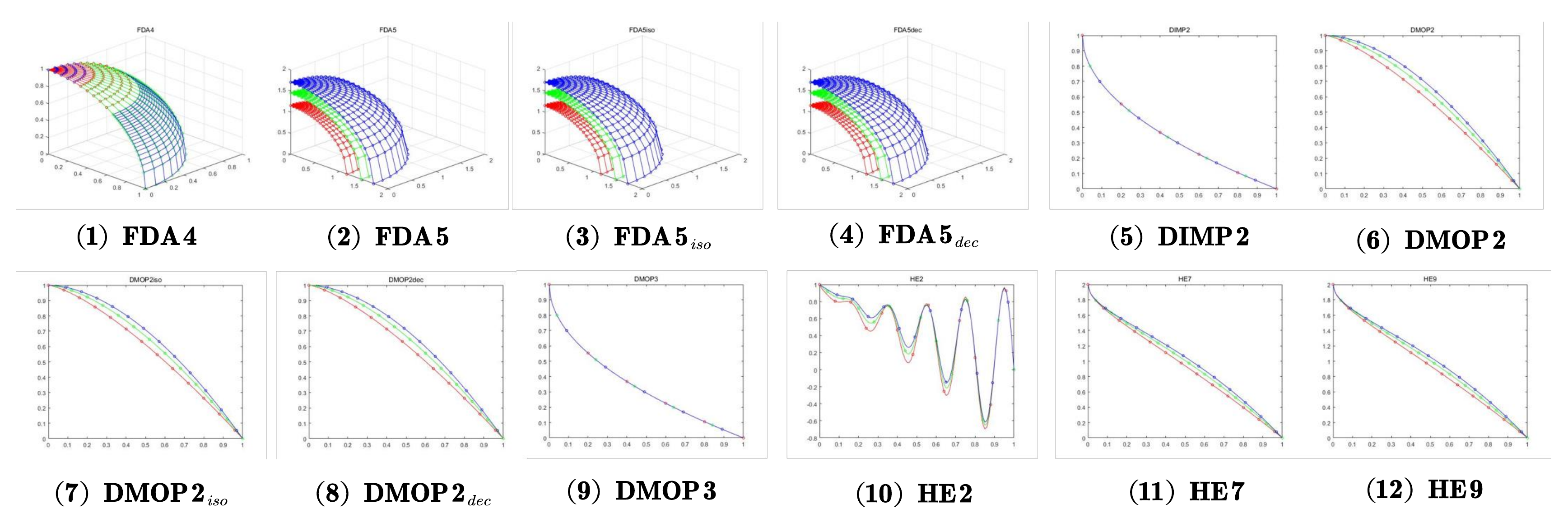}
	\caption{The true POFs of the twelve testing functions. The red, green and blue lines depict the true POFs at the time steps zero, one and two, respectively.}
	\label{Pic:POFs-of-testfuns}
\end{figure*}

The POFs of the  testing functions have different shapes and each function belongs to a certain DMOPs type.  Fig. \ref{Pic:POFs-of-testfuns} describes the true POFs of the twelve testing functions and we let the functions change three times. Type I implies POS changes, but POF does not change; Type II means that the POS and the POF change as well; Type III refers to the condition where the POF changes, but the POS does not change. From Table \ref{Results:testfunctions}, we can find that the POF of the functions could be non-convex, convex, isolated, deceptive, continuous or discontinuous. FDA4 and FDA5 are 3-objective functions while all of the remaining are 2-objective functions.

The dimensions of the decision variables are from 10 to 30-dimension. Please note that the A, B and C values for the functions FDA5$_{iso}$, FDA$_{dec}$, DMOP2$_{iso}$ and DMOP2$_{dec}$ are set to G(t), 0.001 and 0.05 respectively.	

In all of the experiments, we set the population size to 200 and in each generation every algorithm will generate no more than 200 solutions. As mentioned above, we force each benchmark function to change 20 times, and in every change, we let the population carry out 50 iterations.

For the TCA parameters, we set the Gaussian kernel function to the default value and the expected dimensionality was set to be 20. The value of $\mu$ was set to 0.5.

\begin{center}
	\begin{table*}[htbp]
		\renewcommand{\arraystretch}{1.2}
		\begin{minipage}[t]{0.45\textwidth }%
			\centering		
			\caption{The Benchmark Functions}
			\label{Results:testfunctions}
			\resizebox{0.6\textwidth}{!}{
				\begin{tabular}{lccc}\toprule
					\bfseries{Name} & \bfseries{Dim.} & \bfseries{ \# of Obj. } & \bfseries{Type} \\\hline
					FDA4		& 12 & 3 & TYPE I   \\
					FDA5		& 12 & 3 & TYPE II  \\
					FDA5$_{iso}$ 	& 12 & 3 & TYPE II  \\
					FDA5$_{dec}$	& 12 & 3 & TYPE II  \\
					DIMP2		& 10 & 2& TYPE I   \\
					DMOP2		& 10 & 2 & TYPE II  \\
					DMOP2$_{iso}$	& 10 & 2 & TYPE II  \\
					DMOP2$_{dec}$	& 10 & 2 & TYPE II  \\
					DMOP3		& 10 & 2 & TYPE I   \\
					HE2			& 30 & 2 & TYPE III \\
					HE7			& 10 & 2 & TYPE III \\
					HE9			& 10 & 2 & TYPE III \\\bottomrule
				\end{tabular}		
				
			}%
		\end{minipage}
		\begin{minipage}[t]{0.6\textwidth }%
			\centering
			\caption{Configurations of the Benchmark \\Functions Parameters}
			\label{def: different configs}
			\resizebox{0.3\textwidth}{!}{
				\begin{tabular}{cccc}
					\toprule
					& \textbf{$n_t$} & \textbf{$T_t$} & \textbf{$T_T$} \\
					\midrule
					\textbf{C1} & 10    & 5     & 100 \\
					\textbf{C2} & 10    & 10    & 200 \\
					\textbf{C3} & 10    & 25    & 500 \\
					\textbf{C4} & 10    & 50    & 1000 \\
					\textbf{C5} & 1     & 10    & 200 \\
					\textbf{C6} & 1     & 50    & 1000 \\
					\textbf{C7} & 20    & 10    & 200 \\
					\textbf{C8} & 20    & 50    & 1000 \\
					\bottomrule
				\end{tabular}%
				
			}%
		\end{minipage}
	\end{table*}
\end{center}
\subsection{Experimental Results}

\subsubsection{IGD Metric}

For each benchmark function, we perform tests under eight different configurations which are listed in Table \ref{def: different configs}. Each function will change 20 times, and after each change the algorithms would return a POS, and then we calculated the IGD, MIGD, and DMIGD values, respectively.

The detailed results of these experiments are described in twelve tables, and Supplemental Material contains those tables. These tables recorded the MIGD values of the algorithms running on different testing functions under different environments. In these tables, the ``ROC'' refers to the ratio of change of the MIGD values and we used bold face to identify those experiments where performance has been improved.

For the convenience of the readers and at the same time owing to space constraints, we summarize these experimental results in three tables, Table \ref{Results: ROC-Tr-NSGA-II}, Table \ref{Results:ROC-Tr-MOPSO} And Table \ref{Results:ROC-Tr-RM-MEDA}. These three tables illustrate the rate of change of the three new algorithms compared to their original designs. For example, the value of the first row (under C1 configuration) and the first column (under FDA4) of Table \ref{Results: ROC-Tr-NSGA-II} is 61.32, and this shows that the Tr-NSGA-II algorithm improves the NSGA-II algorithm by $61.32\%$ when dealing with the FDA4 problem under the Configuration C1.

\begin{table*}[ht]
	\renewcommand{\arraystretch}{1.25}
	\caption{The Ratio of Change of MIGD Value between Tr-NSGA-II and NSGA-II}
	\label{Results: ROC-Tr-NSGA-II}%
	\centering
	\begin{minipage}[t]{0.75\textwidth}%
		\resizebox{0.99\textwidth}{!}{
			\begin{tabular}{lcccccccc}
				\toprule
				\multicolumn{1}{c}{ROC(\%)} & \multicolumn{1}{c}{C1} & \multicolumn{1}{c}{C2} & \multicolumn{1}{c}{C3} & \multicolumn{1}{c}{C4} & \multicolumn{1}{c}{C5} & \multicolumn{1}{c}{C6} & \multicolumn{1}{c}{C7} & \multicolumn{1}{c}{C8} \\
				\midrule
				FDA4  & \textbf{61.32} & \textbf{62.15} & \textbf{59.42} & \textbf{59.60} & \textbf{78.57} & \textbf{79.75} & \textbf{56.17} & \textbf{61.02} \\
				\midrule
				FDA5  & \textbf{41.13} & \textbf{37.80} & -14.36 & \textbf{14.45} & \textbf{37.12} & \textbf{39.04} & \textbf{20.39} & \textbf{33.09} \\
				\midrule
				FDA5$_{iso}$  & -48.91 & -17.37 & -31.83 & -31.98 & -20.14 & -28.66 & -2.44 & -6.15 \\
				\midrule
				FDA5$_{dec}$  & \textbf{19.83} & \textbf{17.09} & \textbf{26.06} & \textbf{40.23} & \textbf{22.52} & \textbf{3.07} & \textbf{51.88} & \textbf{41.34} \\
				\midrule
				DIMP2  & \textbf{30.70} & \textbf{42.41} & \textbf{39.57} & \textbf{34.61} & \textbf{47.30} & \textbf{38.69} & \textbf{37.57} & \textbf{44.54} \\
				\midrule
				DMOP2  & \textbf{74.97} & \textbf{66.52} & \textbf{78.43} & \textbf{67.01} & -676.29 & \textbf{88.96} & \textbf{61.22} & \textbf{72.65} \\
				\midrule
				DMOP2$_{iso}$  & -4.84 & -4.46 & -6.81 & -3.18 & -7.54 & -13.79 & -2.98 & -2.48 \\
				\midrule
				DMOP2$_{dec}$  & \textbf{44.91} & \textbf{51.25} & \textbf{44.39} & \textbf{40.13} & \textbf{23.26} & \textbf{42.45} & \textbf{45.66} & \textbf{25.58} \\
				\midrule
				DMOP3  & \textbf{72.70} & \textbf{73.61} & \textbf{83.50} & \textbf{57.65} & -514.50 & \textbf{4.32} & \textbf{74.18} & \textbf{76.29} \\
				\midrule
				HE2  & \textbf{2.73} & -17.45 & \textbf{1.67} & \textbf{5.86} & \textbf{67.45} & \textbf{77.01} & \textbf{28.07} & -19.57 \\
				\midrule
				HE7  & \textbf{57.10} & \textbf{58.57} & \textbf{59.45} & \textbf{61.57} & \textbf{53.47} & \textbf{58.34} & \textbf{57.77} & \textbf{61.98} \\
				\midrule
				HE9  & \textbf{14.79} & \textbf{13.02} & \textbf{16.64} & \textbf{12.99} & \textbf{14.98} & \textbf{14.61} & \textbf{16.14} & \textbf{15.49} \\
				\midrule
				\bottomrule
			\end{tabular}%
		}%
	\end{minipage}
\end{table*}

\begin{table*}[ht]
	\renewcommand{\arraystretch}{1.25}
	\caption{The Ratio of Change of MIGD Value between Tr-MOPSO and MOPSO}
	\label{Results:ROC-Tr-MOPSO}%
	\centering
	\begin{minipage}[t]{0.75\textwidth}%
		\resizebox{0.99\textwidth}{!}{
			\begin{tabular}{lcccccccc}
				\toprule
				\multicolumn{1}{c}{ROC(\%)} & \multicolumn{1}{c}{C1} & \multicolumn{1}{c}{C2} & \multicolumn{1}{c}{C3} & \multicolumn{1}{c}{C4} & \multicolumn{1}{c}{C5} & \multicolumn{1}{c}{C6} & \multicolumn{1}{c}{C7} & \multicolumn{1}{c}{C8} \\
				\midrule
				FDA4  & \textbf{18.01} & \textbf{16.44} & \textbf{15.85} & \textbf{17.34} & \textbf{14.15} & \textbf{13.09} & \textbf{18.98} & \textbf{18.88} \\
				\midrule
				FDA5  & \textbf{43.09} & \textbf{56.21} & \textbf{25.44} & \textbf{21.56} & \textbf{60.00} & \textbf{57.88} & \textbf{29.65} & \textbf{23.78} \\
				\midrule
				FDA5$_{iso}$  & -11.03 & -7.58 & -2.12 & -1.49 & -12.51 & -5.11 & -6.98 & -8.00 \\
				\midrule
				FDA5$_{dec}$  & \textbf{17.54} & \textbf{30.66} & \textbf{50.61} & \textbf{23.97} & \textbf{10.64} & \textbf{11.78} & \textbf{28.18} & \textbf{20.68} \\
				\midrule
				DIMP2  & \textbf{86.54} & \textbf{88.36} & \textbf{93.05} & \textbf{94.73} & \textbf{73.76} & \textbf{87.17} & \textbf{83.68} & \textbf{89.37} \\
				\midrule
				DMOP2  & \textbf{92.85} & \textbf{96.77} & \textbf{96.51} & \textbf{94.93} & -478.53 & \textbf{78.24} & \textbf{26.14} & \textbf{84.36} \\
				\midrule
				DMOP2$_{iso}$  & -0.26 & \textbf{2.55} & \textbf{1.64} & \textbf{2.82} & -0.01 & -0.04 & -0.58 & \textbf{3.31} \\
				\midrule
				DMOP2$_{dec}$  & \textbf{89.96} & \textbf{93.98} & \textbf{93.92} & \textbf{96.53} & \textbf{92.54} & \textbf{92.10} & \textbf{97.14} & \textbf{96.66} \\
				\midrule
				DMOP3  & -296.04 & \textbf{54.28} & \textbf{19.34} & \textbf{28.14} & -1371.99 & \textbf{94.39} & \textbf{51.00} & \textbf{26.72} \\
				\midrule
				HE2  & \textbf{14.04} & \textbf{34.03} & \textbf{11.11} & \textbf{34.34} & \textbf{38.46} & \textbf{36.01} & -1.96 & \textbf{14.06} \\
				\midrule
				HE7  & \textbf{2.88} & -3.40 & -1.78 & -0.49 & \textbf{13.96} & \textbf{8.11} & -14.43 & -9.12 \\
				\midrule
				HE9  & -11.81 & -11.45 & -12.29 & -14.04 & -37.00 & -33.30 & -16.15 & -13.57 \\
				\midrule
				\bottomrule
			\end{tabular}%
		}%
	\end{minipage}
\end{table*}

\begin{table*}[ht]
	\renewcommand{\arraystretch}{1.25}
	\caption{The Ratio of Change MIGD Value between Tr-RM-MEDA and RM-MEDA}
	\label{Results:ROC-Tr-RM-MEDA}%
	\centering
	\begin{minipage}[t]{0.75\textwidth}%
		\resizebox{0.99\textwidth}{!}{
			\begin{tabular}{lcccccccc}
				\toprule
				\multicolumn{1}{c}{ROC(\%)} & \multicolumn{1}{c}{C1} & \multicolumn{1}{c}{C2} & \multicolumn{1}{c}{C3} & \multicolumn{1}{c}{C4} & \multicolumn{1}{c}{C5} & \multicolumn{1}{c}{C6} & \multicolumn{1}{c}{C7} & \multicolumn{1}{c}{C8} \\
				\midrule
				FDA4  & \textbf{21.47} & \textbf{22.74} & \textbf{22.42} & \textbf{22.91} & \textbf{25.75} & \textbf{27.71} & \textbf{22.32} & \textbf{23.08} \\
				\midrule
				FDA5  & \textbf{54.07} & \textbf{59.96} & \textbf{61.66} & \textbf{57.40} & \textbf{76.55} & \textbf{69.49} & \textbf{56.24} & \textbf{54.71} \\
				\midrule
				FDA5$_{iso}$  & -0.29 & -0.09 & 0.08  & -0.30 & 1.81  & -0.06 & -0.63 & -0.08 \\
				\midrule
				FDA5$_{dec}$  & \textbf{39.69} & \textbf{33.15} & \textbf{37.16} & \textbf{35.83} & \textbf{65.97} & \textbf{65.59} & \textbf{41.72} & \textbf{45.66} \\
				\midrule
				DIMP2  & \textbf{5.91} & -8.53 & -2.03 & -3.49 & -1.08 & \textbf{1.01} & -2.33 & -4.57 \\
				\midrule
				DMOP2  & -827.79 & -52.71 & \textbf{37.53} & \textbf{33.36} & -3.34 & -1.67 & \textbf{21.07} & -13.04 \\
				\midrule
				DMOP2$_{iso}$  & -0.01 & -0.09 & -0.05 & -0.06 & \textbf{0.01} & -0.01 & \textbf{0.07} & -0.09 \\
				\midrule
				DMOP2$_{dec}$ & \textbf{53.72} & \textbf{64.32} & \textbf{51.05} & \textbf{60.02} & \textbf{1.39} & \textbf{3.85} & \textbf{57.93} & \textbf{49.28} \\
				\midrule
				DMOP3  & \textbf{24.65} & \textbf{22.08} & -25.15 & \textbf{0.49} & -0.45 & -0.79 & \textbf{10.75} & \textbf{27.48} \\
				\midrule
				HE2  & \textbf{86.81} & \textbf{87.01} & \textbf{86.74} & \textbf{88.01} & \textbf{89.99} & \textbf{89.59} & \textbf{88.42} & \textbf{87.91} \\
				\midrule
				HE7  & \textbf{13.31} & \textbf{21.82} & \textbf{21.40} & \textbf{20.29} & \textbf{19.23} & \textbf{14.87} & \textbf{21.32} & \textbf{22.57} \\
				\midrule
				HE9  & \textbf{7.48} & \textbf{7.61} & \textbf{6.95} & \textbf{6.91} & \textbf{8.35} & \textbf{7.53} & \textbf{6.74} & \textbf{5.43} \\
				\midrule
				\bottomrule
			\end{tabular}%
		}%
	\end{minipage}
\end{table*}

Please note that our experimental results are obtained without explicitly tuning the parameters one by one, and if we adjust the parameters separately for different algorithms, we have reason to believe that we can get better experimental results. The reason that we did not tune the parameters specifically for getting better results is that the twelve test functions are not exactly the same, so we can set different parameters to obtain the best performance for each test function. For example, we can construct different latent spaces for the twelve benchmark functions individually via setting different parameters of the TCA method. However, we think that this one-by-one-adjustment strategy does not effectively explain the advantages of our approach since almost all algorithms can obtain a better performance via such parameter-tuning, and this makes no contribution to explain the superiority of the proposed algorithm.

We tabulate all the experimental results and obtain the following observations. For the NSGA-II, the overall effective rate of the Tr-NSGA-II was 78$\%$ (i.e., 75 cases with improving performance out of 96 total tests), of which 33 testing cases increased by more than 50$\%$, 38 increased by 5$\%$ -50$\%$ and 4 cases improved by 0 - 5$\%$; for the MOPSO, the total effective rate was 70$\%$ (i.e., 67 cases with improving performance out of 96 total tests), including 29 testing cases improved by more than 50$\%$, 33 performance improved by 5$\%$ - 50$\%$, and five improved by 0 - 5$\%$; For the RM-MEDA, the total effective rate was 73$\%$ (i.e., 70 cases with improving performance out of 96 total tests), including 23 of the test cases increasing by more than 50$\%$, 39 lifting 5$\%$ to 50$\%$, and 8 improved by  0 - 5$\%$.

These experimental results demonstrate that the transfer learning technique can improve the performance of the existing multiobjective EAs appreciably without significant modifications for solving the DMOPs. On the other hand, we would like to point out that most of the testing cases of performance degradation came from two functions - FDA5$_{iso}$ and DMOP2$_{iso}$. The common characteristic of these two functions is the isolated POFs, so we suspect that the reason why the performance is poor for the two benchmark functions is inappropriate parameters settings.

We also compare the DMIGD value with some chosen state-of-the-art designs, including Multidimensional Bayesian Network based Estimation Distribution Algorithm (MBN-EDA) \cite{karshenas2014multiobjective}, Random immigrants strategy based multiobjective Differential evolutionary algorithm with Decomposition (RND) and the Kalman Filter prediction based DMOEA (MOEA/D-KF) \cite{Muruganantham_2015,Muruganantham_2016}, and the results are depicted in Table \ref{Results:DMIGD}. The experimental results show that the transfer learning based algorithms have much better performance over different problem characteristics in these benchmark functions. Even compared with chosen state-of-the-art algorithms, these transfer learning based algorithms can be much more efficient.

%
%
%

\subsubsection{React Metric}	

Table \ref{Results: DMREACT} depicts the DMReact values of all competing algorithms. We found that the Tr-NSGA-II and Tr-MOPSO have shown improvements, which suggest that, at least at this parameters setting, the proposed approach can improve the adaptability of the NSGA-II and the MOPSO under dynamic environments. However the robustness of the RM-MEDA seems to be reduced, and one reason we envision is that the TCA method have coincidently reduced the diversity of the solutions. So how to improve the diversity and the robustness at the same time is an interesting topic for the future research. The optimal choice of the parameter setting will be a topic in our future research.

\begin{table*}[ht]
	\renewcommand{\arraystretch}{1.25}
	\caption{DMIGD Values of Different Algorithms}
	\label{Results:DMIGD}
	\centering
	\begin{minipage}[t]{0.95\textwidth}%
		\resizebox{0.99\textwidth}{!}{
			\begin{tabular}{lccccccccc}
				\toprule
				\textbf{DMIGD} & {\scriptsize NSGA-II} & {\scriptsize Tr-NSGA-II} & {\scriptsize MOPSO} & {\scriptsize Tr-MOPSO} & {\scriptsize RM-MEDA} & {\scriptsize Tr-RM-MEDA} & {\scriptsize MBN-EDA} & {\scriptsize RND }& {\scriptsize MOEA/D-KF} \\
				\midrule
				FDA4 & 0.2634 & 0.0858 & 0.0732 & 0.0609 & 0.0680 & \textbf{0.0520} & 0.43  & \multicolumn{1}{r}{0.1698} & 0.1913 \\
				\midrule
				FDA5 & 0.3301 & 0.2306 & 0.2131 & 0.1196 & 0.2089 & \textbf{0.0776} & 0.51  & \multicolumn{1}{r}{0.5323} & 0.4963 \\
				\midrule
				FDA5$_{iso}$ & 0.1048 & 0.1292 & 0.1106 & 0.1181 & 0.0650 & \textbf{0.0649} & 0.64  & 0.1433   & 0.1465 \\
				\midrule
				FDA5$_{dec}$ & 0.5923 & 0.4559 & 0.2746 & 0.2139 & 0.5779 & \textbf{0.3275} & 1.27  & 0.5403   & 0.5476 \\
				\midrule
				DIMP2 & 3.8986 & 2.3502 & 2.3684 & \textbf{0.2937} & 4.8892 & 4.9769 & 6.97  & 17.9537   & 22.9536 \\
				\midrule
				DMOP2 & 0.4202 & 0.3439 & \textbf{0.2129} & 0.2538 & 4.5942 & 4.7130 & 1.4   & 1.4329 & 3.0619 \\
				\midrule
				DMOP2$_{iso}$ & 0.0325 & 0.0358 & 0.0319 & 0.0318 & \textbf{0.0290} & \textbf{0.0290} & 2.56  & 0.0315 & 0.0316 \\
				\midrule
				DMOP2$_{dec}$ & 0.6303 & 0.3930 & 0.4192 & \textbf{0.0254} & 0.1449 & 0.0940 & 2.89  & 9.0504 & 9.2188 \\
				\midrule
				DMOP3 & 0.8851 & 1.0133 & 0.2851 & \textbf{0.2650} & 4.5897 & 4.6177 & 1.38  & 0.0697 & 0.0836 \\
				\midrule
				HE2 & 0.2096 & 0.1501 & 0.0847 & \textbf{0.0640} & 0.8451 & 0.1017 & 0.83  & 0.0744 & 0.0745 \\
				\midrule
				HE7 & 0.0946 & 0.0390 & 0.0582 & 0.0583 & 0.0428 & \textbf{0.0342} & 0.21  & 0.1787 & 0.2365 \\
				\midrule
				HE9 & 0.2954 & 0.2508 & 0.2459 & 0.2887 & 0.2565 & \textbf{0.2383} & 0.36  & 0.3432 & 0.4108 \\
				\midrule
				\bottomrule
			\end{tabular}%

		}%
	\end{minipage}
\end{table*}

\begin{table*}[ht]
	\renewcommand{\arraystretch}{1.25}
	\caption{DMReact Value of the Algorithms}
	\label{Results: DMREACT}%
	\centering
	\begin{minipage}[t]{0.75\textwidth}%
		\resizebox{0.99\textwidth}{!}{
			\begin{tabular}{lcccccc}
				\toprule
				\multicolumn{1}{c}{\textbf{DMREACT}} & NSGA-II & Tr-NSGA-II & MOPSO & Tr-MOPSO & RM-MEDA & Tr-RM-MEDA \\
				\midrule
				FDA4  & 1.9803 & \textbf{1.7664} & 1.4934 & \textbf{1.2895} & 1.5033 & 1.7072 \\
				\midrule
				FDA5  & 1.7204 & \textbf{1.5625} & 1.4375 & \textbf{1.3092} & 2.9638 & \textbf{1.6086} \\
				\midrule
				FDA5$_{iso}$ & 1.7105 & \textbf{1.4539} & 1.6283 & 1.7039 & 1.0329 & \textbf{1.0066} \\
				\midrule
				FDA5$_{dec}$ & 1.7928 & 1.8224 & 1.5132 & 1.9375 & 2.5000 & 2.5263 \\
				\midrule
				DIMP2 & 2.2697 & 2.2763 & 1.4013 & \textbf{1.1151} & 1.9243 & 2.0197 \\
				\midrule
				DMOP2 & 2.1645 & \textbf{1.9671} & 1.5592 & 1.8487 & 1.5789 & 1.7467 \\
				\midrule
				DMOP2$_{iso}$ & 1.4375 & 1.4836 & 1.4704 & \textbf{1.3586} & 1.3355 & 1.4605 \\
				\midrule
				DMOP2$_{dec}$ & 2.2961 & \textbf{2.0164} & 1.5461 & 2.1809 & 1.9309 & 2.1316 \\
				\midrule
				DMOP3 & 1.7039 & \textbf{1.3816} & 1.5329 & \textbf{1.3026} & 1.0987 & 1.1349 \\
				\midrule
				HE2   & 2.0428 & \textbf{1.9474} & 1.0987 & 1.2862 & 2.1086 & \textbf{1.4572} \\
				\midrule
				HE7   & 1.0000 & \textbf{1.0000} & 1.0000 & \textbf{1.0000} & 1.0000 & \textbf{1.0000} \\
				\midrule
				HE9   & 1.0000 & \textbf{1.0000} & 1.0000 & 1.3125 & 1.0000 & \textbf{1.0000} \\
				\midrule
				\bottomrule
			\end{tabular}%
		}%
	\end{minipage}
\end{table*}

\section{Conclusion and Future Works}
\label{sec:conclusion}
In this paper, we propose an approach of exploiting a transfer learning technique to enhance the performance of dynamic multiobjective evolutionary algorithms. Our idea is that the solutions of a given DMOP at different times have different distributions, though there are some relationships between these probability distributions, they are not identical. This is a typical non-independent identically distributed (Non-IID) problem, and classical machine learning methods are difficult to solve it.

For this reason, it is not surprise to understand why the traditional dynamic optimization algorithms designed based on classical machine learning find it hard to achieve satisfactory performance. To overcome these problems, we employ the techniques from the transfer learning to develop an algorithmic framework, which creates benefits for a variety of population-based dynamic multiobjective evolutionary algorithms.

In our approach, we consider different probability distributions that the solutions obey at various times as the source and target domains, respectively. We can exploit the gained POF from the source domain to improve computational efficiency in searching for the POF at the next time instance. To achieve this goal, the transfer learning technique is applied to find a latent space where the global feature, MMD value, of the source and target domains is as small as possible. Meanwhile the major statistical characteristics of the data, i.e., the variance will remain unchanged. In this way, we can use the obtained POS to construct an initial population which can be employed by any population-based optimization algorithms to find the POS of the next time instance.

We applied the proposed idea to improve three well-known multidobjective optimization algorithms: NSGA-II, MOPSO, and RM-MEDA. The enhanced algorithms are compared with the original designs and some chosen competing algorithms on a well-adopted benchmark set which involves twelve testing functions. Almost all the experimental results validate that introducing the transfer leaning technique into the dynamic optimization algorithm can greatly improve the quality of the solutions and robustness of the algorithms. This line of research proposed herein can be regarded as a new avenue for designing effective and efficient evolutionary algorithms for DMOPs. A rich body of machine learning techniques can inspire further innovations in solving real-world application \cite{chao2014infant,jiang2010embodied,chao2014robotic,jiang2012fuzzy} with various degrees of complexities and uncertainties.


\section*{Acknowledgments}
This work was supported by the National Natural Science Foundation of China (No.61003014 and No.61673328) and Foundation of Xiamen University's President (No. 20720150150). The first author also wants to thank China Scholarship Council (No. 20150631505) and Oklahoma State University for providing funding and facilities to support his research as a visiting scholar. We are very grateful to Miss Arrchana Muruganantham for providing the code of \cite{Muruganantham_2015} and Mr. Minmin Wang for experimental assistance.

\bibliography{mybibfile}

\begin{thebibliography}{10}

\bibitem{Azzouz_2015}
R.~Azzouz, S.~Bechikh, and L.~B. Said.
\newblock {A dynamic multi-objective evolutionary algorithm using a change
  severity-based adaptive population management strategy}.
\newblock {\em Soft Comput}, aug 2015.

\bibitem{Azzouz_2016}
R.~Azzouz, S.~Bechikh, and L.~B. Said.
\newblock {Dynamic multi-objective optimization using evolutionary algorithms:
  a survey}.
\newblock In {\em Recent Advances in Evolutionary Multi-objective
  Optimization}, pages 31--70. Springer Nature, aug 2016.

\bibitem{BenDavid.2010}
S.~Ben-David, J.~Blitzer, K.~Crammer, A.~Kulesza, F.~Pereira, and J.~W.
  Vaughan.
\newblock A theory of learning from different domains.
\newblock {\em Machine learning}, 79(1-2):151--175, 2010.

\bibitem{BenDavid.2007}
S.~Ben-David, J.~Blitzer, K.~Crammer, F.~Pereira, et~al.
\newblock Analysis of representations for domain adaptation.
\newblock {\em Advances in neural information processing systems}, 19:137,
  2007.

\bibitem{Bosman_2007}
P.~A.~N. Bosman.
\newblock {Learning and anticipation in online dynamic optimization}.
\newblock In {\em Studies in Computational Intelligence}, pages 129--152.
  Springer Science mathplus Business Media, 2007.

\bibitem{Branke}
J.~Branke.
\newblock {Memory enhanced evolutionary algorithms for changing optimization
  problems}.
\newblock In {\em Proceedings of the 1999 Congress on Evolutionary
  Computation}. Institute of Electrical and Electronics Engineers ({IEEE}).

\bibitem{Branke_2000}
J.~Branke, T.~Kaussler, C.~Smidt, and H.~Schmeck.
\newblock {A Multi-population approach to dynamic optimization problems}.
\newblock In {\em Evolutionary Design and Manufacture}, pages 299--307.
  Springer Science mathplus Business Media, 2000.

\bibitem{chao2014robotic}
F.~Chao, F.~Chen, Y.~Shen, W.~He, Y.~Sun, Z.~Wang, C.~Zhou, and M.~Jiang.
\newblock Robotic free writing of chinese characters via human--robot
  interactions.
\newblock {\em International Journal of Humanoid Robotics}, 11(01):1450007,
  2014.

\bibitem{chao2014infant}
F.~Chao, M.~H. Lee, M.~Jiang, and C.~Zhou.
\newblock An infant development-inspired approach to robot hand-eye
  coordination.
\newblock {\em International Journal of Advanced Robotic Systems}, 11, 2014.

\bibitem{cobb1990investigation}
H.~G. Cobb.
\newblock {An investigation into the use of hypermutation as an adaptive
  operator in genetic algorithms having continuous, time-dependent
  nonstationary environments}.
\newblock Technical Report Technical Report AIC-90-001, 1990.

\bibitem{coello2002mopso}
C.~C. Coello and M.~S. Lechuga.
\newblock Mopso: A proposal for multiple objective particle swarm optimization.
\newblock In {\em Evolutionary Computation, 2002. CEC'02. Proceedings of the
  2002 Congress on}, volume~2, pages 1051--1056. IEEE, 2002.

\bibitem{Cruz_2010}
C.~Cruz, J.~R. Gonz{\'{a}}lez, and D.~A. Pelta.
\newblock {Optimization in dynamic environments: a survey on problems methods
  and measures}.
\newblock {\em Soft Comput}, 15(7):1427--1448, dec 2010.

\bibitem{daneshyari2012cultural}
M.~Daneshyari and G.~G. Yen.
\newblock Cultural-based particle swarm for dynamic optimisation problems.
\newblock {\em International Journal of Systems Science}, 43(7):1284--1304,
  2012.

\bibitem{dang2016populations}
D.~Dang, T.~Jansen, and P.~K. Lehre.
\newblock Populations can be essential in tracking dynamic optima.
\newblock {\em Algorithmica}, pages 1--21, 2016.

\bibitem{Dang_2015}
D.-C. Dang, T.~Jansen, and P.~K. Lehre.
\newblock {Populations can be essential in dynamic optimisation}.
\newblock In {\em Proceedings of the 2015 on Genetic and Evolutionary
  Computation Conference - {GECCO},}. Association for Computing Machinery
  ({ACM}), 2015.

\bibitem{Deb}
K.~Deb, U.~B.~R. N., and S.~Karthik.
\newblock {Dynamic multi-objective optimization and decision-Making using
  modified {NSGA}-{II}: a case study on hydro-thermal power scheduling}.
\newblock In {\em Lecture Notes in Computer Science}, pages 803--817. Springer
  Science mathplus Business Media.

\bibitem{deb2002fast}
K.~Deb, A.~Pratap, S.~Agarwal, and T.~Meyarivan.
\newblock A fast and elitist multiobjective genetic algorithm: Nsga-ii.
\newblock {\em IEEE transactions on evolutionary computation}, 6(2):182--197,
  2002.

\bibitem{Farina_2004}
M.~Farina, K.~Deb, and P.~Amato.
\newblock {Dynamic multiobjective optimization problems: test cases
  approximations, and applications}.
\newblock {\em {{{{{{{{{{{{{{{{{{{{{{IEEE} Transactions on Evolutionary
  Computation}}}}}}}}}}}}}}}}}}}}}}, 8(5):425--442, oct 2004.

\bibitem{feng2015memetic}
L.~Feng, Y.-S. Ong, M.-H. Lim, and I.~W. Tsang.
\newblock Memetic search with interdomain learning: A realization between cvrp
  and carp.
\newblock {\em IEEE Transactions on Evolutionary Computation}, 19(5):644--658,
  2015.

\bibitem{Goh}
C.-K. Goh and K.~C. Tan.
\newblock {Dynamic evolutionary multi-objective optimization}.
\newblock In {\em Evolutionary Multi-objective Optimization in Uncertain
  Environments}, pages 125--152. Springer Science mathplus Business Media.

\bibitem{Chi_Keong_Goh_2009}
C.-K. Goh and K.~C. Tan.
\newblock {A competitive-cooperative coevolutionary paradigm for dynamic
  multiobjective optimization}.
\newblock {\em {{{{{{{{{{IEEE} Transactions on Evolutionary
  Computation}}}}}}}}}}, 13(1):103--127, feb 2009.

\bibitem{Grefenstette92geneticalgorithms}
J.~Grefenstette.
\newblock {Genetic algorithms for changing environments}.
\newblock In {\em Parallel Problem Solving from Nature 2}, pages 137--144.
  Elsevier, 1992.

\bibitem{Gretton.2006}
A.~Gretton, K.~M. Borgwardt, M.~Rasch, B.~Sch{\"o}lkopf, and A.~J. Smola.
\newblock A kernel method for the two-sample-problem.
\newblock In {\em Advances in neural information processing systems}, pages
  513--520, 2006.

\bibitem{helbig2015benchmark}
M.~Helbig and A.~Engelbrecht.
\newblock Benchmark functions for cec 2015 special session and competition on
  dynamic multi-objective optimization.
\newblock Technical report, 2015.

\bibitem{iqbal2014reusing}
M.~Iqbal, W.~N. Browne, and M.~Zhang.
\newblock Reusing building blocks of extracted knowledge to solve complex,
  large-scale boolean problems.
\newblock {\em IEEE Transactions on Evolutionary Computation}, 18(4):465--480,
  2014.

\bibitem{iqbal2017cross}
M.~Iqbal, B.~Xue, H.~Al-Sahaf, and M.~Zhang.
\newblock Cross-domain reuse of extracted knowledge in genetic programming for
  image classification.
\newblock {\em IEEE Transactions on Evolutionary Computation}, to be published.

\bibitem{jiang2017integration}
M.~Jiang, W.~Huang, Z.~Huang, and G.~G. Yen.
\newblock Integration of global and local metrics for domain adaptation
  learning via dimensionality reduction.
\newblock {\em IEEE Transactions on Cybernetics}, 47(1):38--51, 2017.

\bibitem{jiang2012fuzzy}
M.~Jiang, Y.~Yu, X.~Liu, F.~Zhang, and Q.~Hong.
\newblock Fuzzy neural network based dynamic path planning.
\newblock In {\em 2012 International Conference on Machine Learning and
  Cybernetics}, volume~1, pages 326--330. IEEE, 2012.

\bibitem{jiang2010embodied}
M.~Jiang, C.~Zhou, and S.~Chen.
\newblock Embodied concept formation and reasoning via neural-symbolic
  integration.
\newblock {\em Neurocomputing}, 74(1):113--120, 2010.

\bibitem{karshenas2014multiobjective}
H.~Karshenas, R.~Santana, C.~Bielza, and P.~Larranaga.
\newblock Multiobjective estimation of distribution algorithm based on joint
  modeling of objectives and variables.
\newblock {\em IEEE Transactions on Evolutionary Computation}, 18(4):519--542,
  2014.

\bibitem{Li_2008}
C.~Li and S.~Yang.
\newblock {Fast Multi-Swarm Optimization for Dynamic Optimization Problems}.
\newblock In {\em 2008 Fourth International Conference on Natural Computation}.
  Institute of Electrical and Electronics Engineers ({IEEE}), 2008.

\bibitem{Mavrovouniotis_2013}
M.~Mavrovouniotis and S.~Yang.
\newblock {Genetic algorithms with adaptive immigrants for dynamic
  environments}.
\newblock In {\em 2013 {IEEE} Congress on Evolutionary Computation}. Institute
  of Electrical {\&} Electronics Engineers ({IEEE}), jun 2013.

\bibitem{Muruganantham_2016}
A.~Muruganantham, K.~Tan, and P.~Vadakkepat.
\newblock {Evolutionary Dynamic Multiobjective Optimization Via Kalman Filter
  Prediction}.
\newblock {\em IEEE Transactions on Cybernetics}, 46(12):2862--2873, Dec 2016.

\bibitem{Muruganantham_2015}
A.~Muruganantham, K.~C. Tan, and P.~Vadakkepat.
\newblock {Solving the {IEEE} {CEC} 2015 dynamic benchmark problems using
  Kalman filter based dynamic multiobjective evolutionary algorithm}.
\newblock In {\em Proceedings in Adaptation Learning and Optimization}, pages
  239--252. Springer Science mathplus Business Media, nov 2015.

\bibitem{nebro2008abyss}
A.~J. Nebro, F.~Luna, E.~Alba, B.~Dorronsoro, J.~J. Durillo, and A.~Beham.
\newblock Abyss: Adapting scatter search to multiobjective optimization.
\newblock {\em IEEE Transactions on Evolutionary Computation}, 12(4):439--457,
  2008.

\bibitem{Nguyen_2012}
T.~T. Nguyen, S.~Yang, and J.~Branke.
\newblock {Evolutionary dynamic optimization: A survey of the state of the
  art}.
\newblock {\em Swarm and Evolutionary Computation}, 6:1--24, oct 2012.

\bibitem{pan2011domain}
S.~J. Pan, I.~W. Tsang, J.~T. Kwok, and Q.~Yang.
\newblock Domain adaptation via transfer component analysis.
\newblock {\em IEEE Transactions on Neural Networks}, 22(2):199--210, 2011.

\bibitem{Pan.2010}
S.~J. Pan and Q.~Yang.
\newblock A survey on transfer learning.
\newblock {\em IEEE Transactions on, Knowledge and Data Engineering},
  22(10):1345--1359, 2010.

\bibitem{Patel.2014}
V.~M. Patel, R.~Gopalan, R.~Li, and R.~Chellappa.
\newblock Visual domain adaptation: an overview of recent advances.
\newblock {\em IEEE signal processing magazine}, 32:53--69, 2015.

\bibitem{Raquel_2013}
C.~Raquel and X.~Yao.
\newblock {Dynamic multi-objective optimization: a survey of the
  state-of-the-art}.
\newblock In {\em Studies in Computational Intelligence}, pages 85--106.
  Springer Science mathplus Business Media, 2013.

\bibitem{Rossi_2008}
C.~Rossi, M.~Abderrahim, and J.~C. D{\'{\i}}az.
\newblock {Tracking moving optima using Kalman-based predictions}.
\newblock {\em Evolutionary Computation}, 16(1):1--30, mar 2008.

\bibitem{Scholkopft.1999}
B.~Scholkopft and K.-R. Mullert.
\newblock Fisher discriminant analysis with kernels.
\newblock In {\em Proceedings of the 1999 IEEE Signal Processing Society
  Workshop Neural Networks for Signal Processing IX, Madison, WI, USA}, pages
  23--25, 1999.

\bibitem{shawe2004kernel}
J.~Shawe-Taylor and N.~Cristianini.
\newblock {\em Kernel methods for pattern analysis}.
\newblock Cambridge university press, 2004.

\bibitem{sierra2005improving}
M.~R. Sierra and C.~A.~C. Coello.
\newblock Improving pso-based multi-objective optimization using crowding,
  mutation and $\epsilon$-dominance.
\newblock In {\em International Conference on Evolutionary Multi-Criterion
  Optimization}, pages 505--519. Springer, 2005.

\bibitem{Sim_es_2013}
A.~Sim{\~{o}}es and E.~Costa.
\newblock {Prediction in evolutionary algorithms for dynamic environments}.
\newblock {\em Soft Comput}, 18(8):1471--1497, oct 2013.

\bibitem{Smola.2007}
A.~Smola, A.~Gretton, L.~Song, and B.~Sch{\"o}lkopf.
\newblock A hilbert space embedding for distributions.
\newblock In {\em Algorithmic Learning Theory}, pages 13--31. Springer, 2007.

\bibitem{sola2010parallel}
M.~C. Sola.
\newblock {\em Parallel processing for dynamic multi-objective optimization}.
\newblock PhD thesis, Ph. D. thesis, Universidad de Granada, Dept. of Computer
  Architecture and Computer Technology, Universidad de Granada, Spain, 2010.

\bibitem{Steinwart.2002}
I.~Steinwart.
\newblock On the influence of the kernel on the consistency of support vector
  machines.
\newblock {\em The Journal of Machine Learning Research}, 2:67--93, 2002.

\bibitem{Stroud_2001}
P.~Stroud.
\newblock {Kalman-extended genetic algorithm for search in nonstationary
  environments with noisy fitness evaluations}.
\newblock {\em {IEEE} Transactions on Evolutionary Computation}, 5(1):66--77,
  2001.

\bibitem{Vavak_1996}
F.~Vavak, T.~C. Fogarty, and K.~Jukes.
\newblock {A genetic algorithm with variable range of local search for tracking
  changing environments}.
\newblock In {\em Parallel Problem Solving from Nature {\textemdash} {PPSN}
  {IV}}, pages 376--385. Springer Science mathplus Business Media, 1996.

\bibitem{Wang_2009}
Y.~Wang and B.~Li.
\newblock {Multi-strategy ensemble evolutionary algorithm for dynamic
  multi-objective optimization}.
\newblock {\em Memetic Comp.}, 2(1):3--24, sep 2009.

\bibitem{while2006faster}
L.~While, P.~Hingston, L.~Barone, and S.~Huband.
\newblock A faster algorithm for calculating hypervolume.
\newblock {\em IEEE transactions on evolutionary computation}, 10(1):29--38,
  2006.

\bibitem{Woldesenbet_2009}
Y.~Woldesenbet and G.~Yen.
\newblock {Dynamic evolutionary algorithm with variable relocation}.
\newblock {\em {{{{{{{{{{{{{{{{{{{IEEE} Transactions on Evolutionary
  Computation}}}}}}}}}}}}}}}}}}}, 13(3):500--513, jun 2009.

\bibitem{XuLuhWhiteEtAl2006}
J.~Xu, P.~B. Luh, F.~B. White, E.~Ni, and K.~Kasiviswanathan.
\newblock Power portfolio optimization in deregulated electricity markets with
  risk management.
\newblock {\em IEEE Transactions on Power Systems}, 21(4):1653--1662, 2006.

\bibitem{Yang_2008}
S.~Yang.
\newblock {Genetic algorithms with memory- and elitism-based immigrants in
  dynamic environments}.
\newblock {\em Evolutionary Computation}, 16(3):385--416, Sep 2008.

\bibitem{Yang_2010}
S.~Yang and C.~Li.
\newblock {A clustering particle swarm optimizer for locating and tracking
  multiple optima in dynamic environments}.
\newblock {\em {{{IEEE} Transactions on Evolutionary Computation}}},
  14(6):959--974, dec 2010.

\bibitem{Yang_2007}
S.~Yang and R.~Tin{\'{o}}s.
\newblock {A hybrid immigrants scheme for genetic algorithms in dynamic
  environments}.
\newblock {\em International Journal of Automation and Computing},
  4(3):243--254, jul 2007.

\bibitem{zhang2008rm}
Q.~Zhang, A.~Zhou, and Y.~Jin.
\newblock Rm-meda: A regularity model-based multiobjective estimation of
  distribution algorithm.
\newblock {\em IEEE Transactions on Evolutionary Computation}, 12(1):41--63,
  2008.

\bibitem{Aimin_Zhou_2014}
A.~Zhou, Y.~Jin, and Q.~Zhang.
\newblock {A population prediction strategy for evolutionary dynamic
  multiobjective optimization}.
\newblock {\em IEEE Transactions on Cybernetics}, 44(1):40--53, jan 2014.

\end{thebibliography}

\bibliographystyle{abbrv}
\newpage

\section{supplemental Materials}
This part includes twelve tables. These tables recorded the MIGD values of the six algorithms, NSGA-II, Tr-NSGA-II, MOPSO, Tr-MOPSO, RM-MEDA and Tr- RM-MEDA, running on the twelve testing functions under eight environments, from C1 to C8. In these tables, the ``ROC'' refers to the ratio of change and we used bold face to identify those experiments where performance has been improved.

\begin{table*}[htbp]
	\renewcommand{\arraystretch}{0.5}
	\caption{FDA4}
	\label{Results:FDA4}
	\centering
	\begin{minipage}[t]{0.85\textwidth}%
		\resizebox{0.99\textwidth}{!}{
			\begin{tabular}{c|cccccccccc}
				\toprule
				\toprule
				&       & { {\scriptsize NSGA-II} }   & {\scriptsize Tr-NSGA-II} & \scriptsize{ROC (\%)}   & {\scriptsize MOPSO }    & {\scriptsize Tr-MOPSO } & \scriptsize{ROC (\%)}  & {\scriptsize RM-MEDA }    & Tr-RM-MEDA  & \scriptsize{ROC (\%)} \\
				\midrule
				\multirow{2}[0]{*}{\textbf{C1}} & MIGD  & 0.2276 & 0.0881 & \multirow{2}[0]{*}{\textbf{61.3188} } & 0.0764 & 0.0627 & \multirow{2}[0]{*}{\textbf{18.0104} } & 0.0678 & 0.0533 & \multirow{2}[0]{*}{\textbf{21.4710}} \\
				& VAR   & 1.8246E-05 & 9.2507E-07 &       & 7.7743E-05 & 1.3804E-08 &       & 5.2059E-07 & 3.2465E-07 &  \\
				\midrule
				\multirow{2}[0]{*}{\textbf{C2}} & MIGD  & 0.2244 & 0.0849 & \multirow{2}[0]{*}{\textbf{62.1480} } & 0.0760 & 0.0635 & \multirow{2}[0]{*}{\textbf{16.4435} } & 0.0678 & 0.0524 & \multirow{2}[0]{*}{\textbf{22.7386}} \\
				& VAR   & 4.8821E-05 & 2.8997E-07 &       & 7.7034E-05 & 3.6274E-08 &       & 3.3605E-07 & 4.7269E-07 &  \\
				\midrule
				\multirow{2}[0]{*}{\textbf{C3}} & MIGD  & 0.2123 & 0.0861 & \multirow{2}[0]{*}{\textbf{59.4203} } & 0.0756 & 0.0636 & \multirow{2}[0]{*}{\textbf{15.8521} } & 0.0673 & 0.0522 & \multirow{2}[0]{*}{\textbf{22.4152}} \\
				& VAR   & 6.7454E-05 & 6.7844E-07 &       & 6.4871E-05 & 2.1779E-09 &       & 1.7372E-07 & 1.7532E-07 &  \\
				\midrule
				\multirow{2}[0]{*}{\textbf{C4}} & MIGD  & 0.2121 & 0.0857 & \multirow{2}[0]{*}{\textbf{59.6021} } & 0.0766 & 0.0633 & \multirow{2}[0]{*}{\textbf{17.3353} } & 0.0684 & 0.0527 & \multirow{2}[0]{*}{\textbf{22.9059}} \\
				& VAR   & 1.3541E-05 & 1.2315E-06 &       & 6.4677E-05 & 1.3470E-07 &       & 1.6080E-07 & 1.5401E-07 &  \\
				\midrule
				\multirow{2}[0]{*}{\textbf{C5}} & MIGD  & 0.3982 & 0.0853 & \multirow{2}[0]{*}{\textbf{78.5738} } & 0.0624 & 0.0536 & \multirow{2}[0]{*}{\textbf{14.1528} } & 0.0673 & 0.0500 & \multirow{2}[0]{*}{\textbf{25.7460}} \\
				& VAR   & 4.6983E-04 & 1.9502E-07 &       & 4.8391E-05 & 7.7322E-07 &       & 4.6476E-07 & 5.0149E-11 &  \\
				\midrule
				\multirow{2}[0]{*}{\textbf{C6}} & MIGD  & 0.4151 & 0.0840 & \multirow{2}[0]{*}{\textbf{79.7524} } & 0.0616 & 0.0535 & \multirow{2}[0]{*}{\textbf{13.0936} } & 0.0693 & 0.0501 & \multirow{2}[0]{*}{\textbf{27.7051}} \\
				& VAR   & 5.8715E-05 & 3.7560E-07 &       & 5.8574E-05 & 1.5520E-09 &       & 2.2626E-08 & 6.8992E-08 &  \\
				\midrule
				\multirow{2}[0]{*}{\textbf{C7}} & MIGD  & 0.2003 & 0.0878 & \multirow{2}[0]{*}{\textbf{56.1683} } & 0.0789 & 0.0639 & \multirow{2}[0]{*}{\textbf{18.9788} } & 0.0674 & 0.0523 & \multirow{2}[0]{*}{\textbf{22.3155}} \\
				& VAR   & 2.9321E-10 & 7.0748E-06 &       & 1.3014E-04 & 4.5717E-10 &       & 4.1111E-07 & 8.7830E-08 &  \\
				\midrule
				\multirow{2}[0]{*}{\textbf{C8}} & MIGD  & 0.2174 & 0.0847 & \multirow{2}[0]{*}{\textbf{61.0192} } & 0.0779 & 0.0632 & \multirow{2}[0]{*}{\textbf{18.8763} } & 0.0687 & 0.0529 & \multirow{2}[0]{*}{\textbf{23.0764}} \\
				& VAR   & 1.6242E-04 & 5.8681E-07 &       & 9.0152E-05 & 3.5372E-08 &       & 9.4215E-09 & 3.9562E-09 &  \\
				\bottomrule
			\end{tabular}%
			
		}%
	\end{minipage}\\

	\renewcommand{\arraystretch}{0.5}
	\caption{FDA5}
	\label{Results:FDA5}
	\centering
	\begin{minipage}[t]{0.85\textwidth}%
		\resizebox{0.99\textwidth}{!}{
			\begin{tabular}{c|cccccccccc}
				\toprule
				\toprule
				&       & { {\scriptsize NSGA-II} }   & {\scriptsize Tr-NSGA-II} & \scriptsize{ROC (\%)}   & {\scriptsize MOPSO }    & {\scriptsize Tr-MOPSO } & \scriptsize{ROC (\%)}  & {\scriptsize RM-MEDA }    & Tr-RM-MEDA  & \scriptsize{ROC (\%)} \\
				\midrule
				\multirow{2}[2]{*}{\textbf{C1}} & MIGD  & 0.2400 & 0.1413 & \multirow{2}[2]{*}{\textbf{41.1326 }} & 0.2077 & 0.1182 & \multirow{2}[2]{*}{\textbf{43.0924 }} & 0.1741 & 0.0800 & \multirow{2}[2]{*}{\textbf{54.0663}} \\
				& VAR   & 1.5776E-03 & 5.1236E-06 &       & 4.5041E-03 & 5.1385E-09 &       & 3.1584E-04 & 1.3341E-06 &  \\
				\midrule
				\multirow{2}[2]{*}{\textbf{C2}} & MIGD  & 0.2309 & 0.1437 & \multirow{2}[2]{*}{\textbf{37.7968 }} & 0.2721 & 0.1191 & \multirow{2}[2]{*}{\textbf{56.2091 }} & 0.1973 & 0.0790 & \multirow{2}[2]{*}{\textbf{59.9575}} \\
				& VAR   & 4.9722E-05 & 2.2216E-04 &       & 3.4133E-02 & 3.8575E-06 &       & 1.7169E-03 & 2.6424E-07 &  \\
				\midrule
				\multirow{2}[2]{*}{\textbf{C3}} & MIGD  & 0.2283 & 0.2611 & \multirow{2}[2]{*}{-14.3646 } & 0.1660 & 0.1238 & \multirow{2}[2]{*}{\textbf{25.4400 }} & 0.2091 & 0.0802 & \multirow{2}[2]{*}{\textbf{61.6615}} \\
				& VAR   & 1.9969E-07 & 2.0216E-04 &       & 1.3965E-04 & 2.5556E-06 &       & 1.5582E-03 & 2.8186E-07 &  \\
				\midrule
				\multirow{2}[2]{*}{\textbf{C4}} & MIGD  & 0.2254 & 0.1929 & \multirow{2}[2]{*}{\textbf{14.4507 }} & 0.1545 & 0.1212 & \multirow{2}[2]{*}{\textbf{21.5557 }} & 0.1863 & 0.0793 & \multirow{2}[2]{*}{\textbf{57.4027}} \\
				& VAR   & 3.4071E-04 & 1.9428E-05 &       & 9.5470E-04 & 2.9989E-08 &       & 9.5869E-06 & 7.8247E-08 &  \\
				\midrule
				\multirow{2}[2]{*}{\textbf{C5}} & MIGD  & 0.6507 & 0.4092 & \multirow{2}[2]{*}{\textbf{37.1174 }} & 0.2825 & 0.1130 & \multirow{2}[2]{*}{\textbf{60.0003 }} & 0.2978 & 0.0698 & \multirow{2}[2]{*}{\textbf{76.5526}} \\
				& VAR   & 1.4819E-04 & 1.2949E-03 &       & 1.6887E-04 & 1.7385E-05 &       & 4.1144E-03 & 4.0474E-07 &  \\
				\midrule
				\multirow{2}[2]{*}{\textbf{C6}} & MIGD  & 0.6709 & 0.4090 & \multirow{2}[2]{*}{\textbf{39.0381 }} & 0.3015 & 0.1270 & \multirow{2}[2]{*}{\textbf{57.8789 }} & 0.2682 & 0.0818 & \multirow{2}[2]{*}{\textbf{69.4866}} \\
				& VAR   & 2.0832E-03 & 7.0414E-04 &       & 2.8644E-03 & 1.0997E-03 &       & 2.0990E-03 & 1.6960E-04 &  \\
				\midrule
				\multirow{2}[2]{*}{\textbf{C7}} & MIGD  & 0.1884 & 0.1499 & \multirow{2}[2]{*}{\textbf{20.3940 }} & 0.1654 & 0.1164 & \multirow{2}[2]{*}{\textbf{29.6483 }} & 0.1726 & 0.0755 & \multirow{2}[2]{*}{\textbf{56.2418}} \\
				& VAR   & 2.4646E-05 & 3.2471E-08 &       & 1.0135E-04 & 6.6345E-05 &       & 4.3048E-05 & 6.5608E-10 &  \\
				\midrule
				\multirow{2}[1]{*}{\textbf{C8}} & MIGD  & 0.2057 & 0.1377 & \multirow{2}[1]{*}{\textbf{33.0929 }} & 0.1547 & 0.1179 & \multirow{2}[1]{*}{\textbf{23.7802 }} & 0.1661 & 0.0752 & \multirow{2}[1]{*}{\textbf{54.7071}} \\
				& VAR   & 1.9567E-05 & 5.2939E-04 &       & 1.8381E-05 & 2.1885E-05 &       & 2.7332E-04 & 4.7222E-08 &  \\
				\bottomrule
			\end{tabular}%
			
		}%
	\end{minipage}
\end{table*}

\begin{table*}[htbp]
	\renewcommand{\arraystretch}{0.5}
	\caption{FDA5$_{iso}$}
	\label{Results:FDA5_iso}
	\centering
	\begin{minipage}[t]{0.85\textwidth}%
		\resizebox{0.99\textwidth}{!}{
			
			\begin{tabular}{c|cccccccccc}
				\toprule
				\toprule
				&       & { {\scriptsize NSGA-II} }   & {\scriptsize Tr-NSGA-II} & \scriptsize{ROC (\%)}   & {\scriptsize MOPSO }    & {\scriptsize Tr-MOPSO } & \scriptsize{ROC (\%)}  & {\scriptsize RM-MEDA }    & Tr-RM-MEDA  & \scriptsize{ROC (\%)} \\
				\midrule
				\multirow{2}[2]{*}{\textbf{C1}} & MIGD  & 0.0990 & 0.1474 & \multirow{2}[2]{*}{-48.9093 } & 0.1143 & 0.1270 & \multirow{2}[2]{*}{-11.0275 } & 0.0666 & 0.0668 & \multirow{2}[2]{*}{-0.2884} \\
				& VAR   & 7.4915E-06 & 2.1071E-03 &       & 4.7170E-09 & 1.2904E-04 &       & 3.7265E-09 & 2.6494E-08 &  \\
				\midrule
				\multirow{2}[2]{*}{\textbf{C2}} & MIGD  & 0.1062 & 0.1246 & \multirow{2}[2]{*}{-17.3705 } & 0.1137 & 0.1223 & \multirow{2}[2]{*}{-7.5808 } & 0.0662 & 0.0663 & \multirow{2}[2]{*}{-0.0852} \\
				& VAR   & 1.7928E-04 & 1.5697E-03 &       & 1.9681E-05 & 3.2967E-05 &       & 5.3824E-09 & 3.0749E-08 &  \\
				\midrule
				\multirow{2}[2]{*}{\textbf{C3}} & MIGD  & 0.0999 & 0.1318 & \multirow{2}[2]{*}{-31.8317 } & 0.1169 & 0.1194 & \multirow{2}[2]{*}{-2.1164 } & 0.0669 & 0.0669 & \multirow{2}[2]{*}{\textbf{0.0839}} \\
				& VAR   & 5.3260E-06 & 4.5014E-04 &       & 2.2940E-08 & 4.5508E-06 &       & 1.7117E-07 & 2.6062E-09 &  \\
				\midrule
				\multirow{2}[2]{*}{\textbf{C4}} & MIGD  & 0.0959 & 0.1266 & \multirow{2}[2]{*}{-31.9813 } & 0.1156 & 0.1174 & \multirow{2}[2]{*}{-1.4932 } & 0.0659 & 0.0661 & \multirow{2}[2]{*}{-0.2978} \\
				& VAR   & 3.9586E-06 & 9.6354E-05 &       & 2.3902E-05 & 1.6925E-04 &       & 4.5401E-09 & 1.5496E-08 &  \\
				\midrule
				\multirow{2}[2]{*}{\textbf{C5}} & MIGD  & 0.1245 & 0.1496 & \multirow{2}[2]{*}{-20.1368 } & 0.1026 & 0.1155 & \multirow{2}[2]{*}{-12.5111 } & 0.0624 & 0.0613 & \multirow{2}[2]{*}{\textbf{1.8057}} \\
				& VAR   & 1.0807E-04 & 2.2636E-06 &       & 2.1496E-07 & 7.3640E-05 &       & 4.0145E-06 & 1.0112E-07 &  \\
				\midrule
				\multirow{2}[2]{*}{\textbf{C6}} & MIGD  & 0.1098 & 0.1413 & \multirow{2}[2]{*}{-28.6611 } & 0.1053 & 0.1107 & \multirow{2}[2]{*}{-5.1149 } & 0.0611 & 0.0612 & \multirow{2}[2]{*}{-0.0557} \\
				& VAR   & 3.2348E-06 & 1.4977E-04 &       & 1.6652E-05 & 1.4460E-10 &       & 1.2519E-08 & 2.1923E-09 &  \\
				\midrule
				\multirow{2}[2]{*}{\textbf{C7}} & MIGD  & 0.0950 & 0.0973 & \multirow{2}[2]{*}{-2.4419 } & 0.1062 & 0.1136 & \multirow{2}[2]{*}{-6.9763 } & 0.0651 & 0.0655 & \multirow{2}[2]{*}{-0.6285} \\
				& VAR   & 1.4644E-06 & 1.9496E-05 &       & 2.3870E-06 & 1.1739E-06 &       & 5.1521E-08 & 1.9050E-07 &  \\
				\midrule
				\multirow{2}[2]{*}{\textbf{C8}} & MIGD  & 0.1081 & 0.1148 & \multirow{2}[2]{*}{-6.1528 } & 0.1100 & 0.1188 & \multirow{2}[2]{*}{-8.0000 } & 0.0653 & 0.0654 & \multirow{2}[2]{*}{-0.0824} \\
				& VAR   & 3.9038E-04 & 5.5916E-04 &       & 2.5876E-06 & 9.8240E-05 &       & 2.1896E-09 & 6.6672E-09 &  \\
				\bottomrule
			\end{tabular}%
		}%
	\end{minipage}
	
	\renewcommand{\arraystretch}{0.5}
	\caption{FDA5$_{dec}$}
	\label{Results:FDA5_dec}
	\centering
	\begin{minipage}[t]{0.85\textwidth}%
		\resizebox{0.99\textwidth}{!}{
			\begin{tabular}{c|cccccccccc}
				\toprule
				\toprule
				&       & { {\scriptsize NSGA-II} }   & {\scriptsize Tr-NSGA-II} & \scriptsize{ROC (\%)}   & {\scriptsize MOPSO }    & {\scriptsize Tr-MOPSO } & \scriptsize{ROC (\%)}  & {\scriptsize RM-MEDA }    & Tr-RM-MEDA  & \scriptsize{ROC (\%)} \\
				\midrule
				\textbf{C1} & MIGD  & 0.3701 & 0.2967 & \multirow{2}[2]{*}{\textbf{19.8300 }} & 0.2068 & \multirow{2}[2]{*}{0.1706} & \multirow{2}[2]{*}{\textbf{17.5353 }} & 0.6791 & 0.4096 & \multirow{2}[2]{*}{\textbf{39.6884}} \\
				& VAR   & 1.6658E-03 & 4.1428E-07 &       & 1.1357E-04 &       &       & 6.6137E-04 & 3.0126E-04 &  \\
				\midrule
				\textbf{C2} & MIGD  & 0.3640 & 0.3018 & \multirow{2}[2]{*}{\textbf{17.0947 }} & 0.2113 & \multirow{2}[2]{*}{0.1465} & \multirow{2}[2]{*}{\textbf{30.6647 }} & 0.6170 & 0.4124 & \multirow{2}[2]{*}{\textbf{33.1499}} \\
				& VAR   & 2.5492E-03 & 1.1655E-03 &       & 4.5714E-03 &       &       & 1.6537E-04 & 5.1870E-06 &  \\
				\midrule
				\textbf{C3} & MIGD  & 0.4321 & 0.3194 & \multirow{2}[2]{*}{\textbf{26.0648 }} & 0.2840 & \multirow{2}[2]{*}{0.1403} & \multirow{2}[2]{*}{\textbf{50.6122 }} & 0.6352 & 0.3992 & \multirow{2}[2]{*}{\textbf{37.1567}} \\
				& VAR   & 1.9916E-02 & 8.9818E-06 &       & 4.2669E-05 &       &       & 2.5114E-03 & 4.9007E-04 &  \\
				\midrule
				\textbf{C4} & MIGD  & 0.4584 & 0.2740 & \multirow{2}[2]{*}{\textbf{40.2280 }} & 0.1882 & \multirow{2}[2]{*}{0.1431} & \multirow{2}[2]{*}{\textbf{23.9749 }} & 0.6393 & 0.4102 & \multirow{2}[2]{*}{\textbf{35.8319}} \\
				& VAR   & 1.5253E-03 & 1.6655E-04 &       & 4.4340E-05 &       &       & 5.9788E-04 & 4.2766E-04 &  \\
				\midrule
				\textbf{C5} & MIGD  & 1.1684 & 0.9052 & \multirow{2}[2]{*}{\textbf{22.5197 }} & 0.4569 & \multirow{2}[2]{*}{0.4083} & \multirow{2}[2]{*}{\textbf{10.6388 }} & 0.3726 & 0.1268 & \multirow{2}[2]{*}{\textbf{65.9669}} \\
				& VAR   & 6.3473E-04 & 4.1290E-02 &       & 4.1138E-03 &       &       & 6.1639E-06 & 1.4866E-05 &  \\
				\midrule
				\textbf{C6} & MIGD  & 1.1842 & 1.1479 & \multirow{2}[2]{*}{\textbf{3.0703 }} & 0.4802 & \multirow{2}[2]{*}{0.4237} & \multirow{2}[2]{*}{\textbf{11.7771 }} & 0.3802 & 0.1308 & \multirow{2}[2]{*}{\textbf{65.5900}} \\
				& VAR   & 1.1343E-02 & 4.5164E-02 &       & 4.8785E-03 &       &       & 3.4097E-03 & 7.1229E-04 &  \\
				\midrule
				\textbf{C7} & MIGD  & 0.4179 & 0.2011 & \multirow{2}[2]{*}{\textbf{51.8789 }} & 0.1897 & \multirow{2}[2]{*}{0.1363} & \multirow{2}[2]{*}{\textbf{28.1784 }} & 0.6260 & 0.3648 & \multirow{2}[2]{*}{\textbf{41.7235}} \\
				& VAR   & 4.7462E-03 & 5.6786E-05 &       & 2.8357E-05 &       &       & 1.7293E-04 & 1.9517E-05 &  \\
				\midrule
				\multirow{2}[2]{*}{\textbf{C8}} & MIGD  & 0.3431 & 0.2013 & \multirow{2}[2]{*}{\textbf{41.3357 }} & 0.1794 & \multirow{2}[2]{*}{0.1423} & \multirow{2}[2]{*}{\textbf{20.6783 }} & 0.6739 & 0.3662 & \multirow{2}[2]{*}{\textbf{45.6626}} \\
				& VAR   & 1.8925E-05 & 2.4026E-03 &       & 9.9454E-04 &       &       & 3.7766E-03 & 1.9394E-04 &  \\
				\bottomrule
			\end{tabular}%
			
		}%
	\end{minipage}
\end{table*}

\begin{table*}[htbp]
	\renewcommand{\arraystretch}{0.5}
	\caption{DIMP2}
	\label{Results:DIMP2}
	\centering
	\begin{minipage}[t]{0.85\textwidth}%
		\resizebox{0.99\textwidth}{!}{
			\begin{tabular}{c|cccccccccc}
				\toprule
				\toprule
				&       & { {\scriptsize NSGA-II} }   & {\scriptsize Tr-NSGA-II} & \scriptsize{ROC (\%)}   & {\scriptsize MOPSO }    & {\scriptsize Tr-MOPSO } & \scriptsize{ROC (\%)}  & {\scriptsize RM-MEDA }    & Tr-RM-MEDA  & \scriptsize{ROC (\%)} \\
				\midrule
				\multirow{2}[2]{*}{\textbf{C1}} & MIGD  & 3.6287 & \multirow{2}[2]{*}{2.5147} & \multirow{2}[2]{*}{\textbf{30.6992 }} & 2.9797 & \multirow{2}[2]{*}{0.4010} & \multirow{2}[2]{*}{\textbf{86.5408 }} & 5.2097 & \multirow{2}[2]{*}{4.9018} & \multirow{2}[2]{*}{\textbf{5.9113}} \\
				& VAR   & 1.8027E-01 &       &       & 5.3940E-02 &       &       & 1.7667E-02 &       &  \\
				\midrule
				\multirow{2}[2]{*}{\textbf{C2}} & MIGD  & 4.0771 & \multirow{2}[2]{*}{2.3481} & \multirow{2}[2]{*}{\textbf{42.4074 }} & 2.4547 & \multirow{2}[2]{*}{0.2857} & \multirow{2}[2]{*}{\textbf{88.3599 }} & 4.8282 & \multirow{2}[2]{*}{5.2399} & \multirow{2}[2]{*}{-8.5262} \\
				& VAR   & 4.7095E-02 &       &       & 1.8199E+00 &       &       & 1.1118E-03 &       &  \\
				\midrule
				\multirow{2}[2]{*}{\textbf{C3}} & MIGD  & 4.1530 & \multirow{2}[2]{*}{2.5097} & \multirow{2}[2]{*}{\textbf{39.5693 }} & 2.1233 & \multirow{2}[2]{*}{0.1475} & \multirow{2}[2]{*}{\textbf{93.0529 }} & 4.8860 & \multirow{2}[2]{*}{4.9851} & \multirow{2}[2]{*}{-2.0293} \\
				& VAR   & 1.1714E-01 &       &       & 3.1945E-02 &       &       & 2.2842E-01 &       &  \\
				\midrule
				\multirow{2}[2]{*}{\textbf{C4}} & MIGD  & 3.6302 & \multirow{2}[2]{*}{2.3739} & \multirow{2}[2]{*}{\textbf{34.6066 }} & 2.7988 & \multirow{2}[2]{*}{0.1476} & \multirow{2}[2]{*}{\textbf{94.7278 }} & 4.6559 & \multirow{2}[2]{*}{4.8186} & \multirow{2}[2]{*}{-3.4948} \\
				& VAR   & 1.6550E-01 &       &       & 1.1552E+00 &       &       & 5.6086E-05 &       &  \\
				\midrule
				\multirow{2}[2]{*}{\textbf{C5}} & MIGD  & 4.4689 & \multirow{2}[2]{*}{2.3552} & \multirow{2}[2]{*}{\textbf{47.2980 }} & 1.8512 & \multirow{2}[2]{*}{0.4857} & \multirow{2}[2]{*}{\textbf{73.7642 }} & 4.9163 & \multirow{2}[2]{*}{4.9695} & \multirow{2}[2]{*}{-1.0823} \\
				& VAR   & 1.2814E-01 &       &       & 2.7176E-01 &       &       & 8.7425E-02 &       &  \\
				\midrule
				\multirow{2}[2]{*}{\textbf{C6}} & MIGD  & 3.5897 & \multirow{2}[2]{*}{2.2009} & \multirow{2}[2]{*}{\textbf{38.6900 }} & 2.2850 & \multirow{2}[2]{*}{0.2932} & \multirow{2}[2]{*}{\textbf{87.1703 }} & 5.0086 & \multirow{2}[2]{*}{4.9583} & \multirow{2}[2]{*}{\textbf{1.0050}} \\
				& VAR   & 9.8967E-03 &       &       & 2.5271E-01 &       &       & 2.7665E-01 &       &  \\
				\midrule
				\multirow{2}[2]{*}{\textbf{C7}} & MIGD  & 3.7563 & \multirow{2}[2]{*}{2.3449} & \multirow{2}[2]{*}{\textbf{37.5743 }} & 2.0250 & \multirow{2}[2]{*}{0.3306} & \multirow{2}[2]{*}{\textbf{83.6751 }} & 4.7298 & \multirow{2}[2]{*}{4.8399} & \multirow{2}[2]{*}{-2.3276} \\
				& VAR   & 2.4725E-04 &       &       & 6.1787E-03 &       &       & 1.6563E-03 &       &  \\
				\midrule
				\multirow{2}[2]{*}{\textbf{C8}} & MIGD  & 3.8851 & \multirow{2}[2]{*}{2.1546} & \multirow{2}[2]{*}{\textbf{44.5419 }} & 2.4292 & \multirow{2}[2]{*}{0.2581} & \multirow{2}[2]{*}{\textbf{89.3735 }} & 4.8794 & \multirow{2}[2]{*}{5.1023} & \multirow{2}[2]{*}{-4.5683} \\
				& VAR   & 2.5013E-01 &       &       & 5.2389E-01 &       &       & 3.5292E-02 &       &  \\
				\bottomrule
			\end{tabular}%
			
		}%
	\end{minipage}
	
	\renewcommand{\arraystretch}{0.5}
	\caption{DMOP2}
	\label{Results:DMOP2}
	\centering
	\begin{minipage}[t]{0.85\textwidth}%
		\resizebox{0.99\textwidth}{!}{
			\begin{tabular}{c|cccccccccc}
				\toprule
				\toprule
				&       & { {\scriptsize NSGA-II} }   & {\scriptsize Tr-NSGA-II} & \scriptsize{ROC (\%)}   & {\scriptsize MOPSO }    & {\scriptsize Tr-MOPSO } & \scriptsize{ROC (\%)}  & {\scriptsize RM-MEDA }    & Tr-RM-MEDA  & \scriptsize{ROC (\%)} \\
				\midrule
				\multirow{2}[2]{*}{\textbf{C1}} & MIGD  & 0.1367 & 0.0342 & \multirow{2}[2]{*}{\textbf{74.9665 }} & 0.1472 & 0.0105 & \multirow{2}[2]{*}{\textbf{92.8490 }} & 0.0039 & 0.0366 & \multirow{2}[2]{*}{-827.7905} \\
				& VAR   & 8.3806E-05 & 2.2300E-07 &       & 1.4966E-05 & 6.4208E-06 &       & 8.9592E-08 & 2.9249E-04 &  \\
				\midrule
				\multirow{2}[2]{*}{\textbf{C2}} & MIGD  & 0.1097 & 0.0367 & \multirow{2}[2]{*}{\textbf{66.5239 }} & 0.2784 & 0.0090 & \multirow{2}[2]{*}{\textbf{96.7735 }} & 0.0043 & 0.0066 & \multirow{2}[2]{*}{-52.7081} \\
				& VAR   & 3.4897E-04 & 5.6430E-06 &       & 1.7974E-02 & 9.0819E-07 &       & 1.6877E-06 & 2.7414E-05 &  \\
				\midrule
				\multirow{2}[2]{*}{\textbf{C3}} & MIGD  & 0.1392 & 0.0300 & \multirow{2}[2]{*}{\textbf{78.4328 }} & 0.2780 & 0.0097 & \multirow{2}[2]{*}{\textbf{96.5132 }} & 0.0047 & 0.0029 & \multirow{2}[2]{*}{\textbf{37.5324}} \\
				& VAR   & 1.5812E-03 & 2.4597E-07 &       & 2.3922E-05 & 2.0104E-06 &       & 4.4530E-07 & 4.7458E-08 &  \\
				\midrule
				\multirow{2}[2]{*}{\textbf{C4}} & MIGD  & 0.1234 & 0.0407 & \multirow{2}[2]{*}{\textbf{67.0124 }} & 0.1821 & 0.0092 & \multirow{2}[2]{*}{\textbf{94.9274 }} & 0.0041 & 0.0027 & \multirow{2}[2]{*}{\textbf{33.3552}} \\
				& VAR   & 3.4172E-04 & 1.5152E-04 &       & 8.6546E-03 & 1.1422E-07 &       & 1.1866E-07 & 2.5738E-08 &  \\
				\midrule
				\multirow{2}[2]{*}{\textbf{C5}} & MIGD  & 0.2939 & 2.2819 & \multirow{2}[2]{*}{-676.2864 } & 0.3241 & 1.8752 & \multirow{2}[2]{*}{-478.5327 } & 18.3749 & 18.9887 & \multirow{2}[2]{*}{-3.3402} \\
				& VAR   & 1.5190E-04 & 8.0239E+00 &       & 1.1403E-01 & 6.3602E+00 &       & 9.8567E-03 & 1.3102E-01 &  \\
				\midrule
				\multirow{2}[2]{*}{\textbf{C6}} & MIGD  & 2.3484 & 0.2593 & \multirow{2}[2]{*}{\textbf{88.9568 }} & 0.4015 & 0.0874 & \multirow{2}[2]{*}{\textbf{78.2424 }} & 18.3544 & 18.6601 & \multirow{2}[2]{*}{-1.6652} \\
				& VAR   & 7.5218E+00 & 1.3435E-03 &       & 1.8818E-02 & 8.9815E-06 &       & 5.7303E-03 & 3.1135E-02 &  \\
				\midrule
				\multirow{2}[2]{*}{\textbf{C7}} & MIGD  & 0.0967 & 0.0375 & \multirow{2}[2]{*}{\textbf{61.2215 }} & 0.0252 & 0.0186 & \multirow{2}[2]{*}{\textbf{26.1430 }} & 0.0035 & 0.0028 & \multirow{2}[2]{*}{\textbf{21.0670}} \\
				& VAR   & 5.6724E-05 & 9.6483E-05 &       & 6.0961E-08 & 2.2825E-04 &       & 6.1515E-08 & 7.2391E-08 &  \\
				\midrule
				\multirow{2}[2]{*}{\textbf{C8}} & MIGD  & 0.1136 & 0.0311 & \multirow{2}[2]{*}{\textbf{72.6521 }} & 0.0668 & 0.0104 & \multirow{2}[2]{*}{\textbf{84.3618 }} & 0.0034 & 0.0038 & \multirow{2}[2]{*}{-13.0372} \\
				& VAR   & 8.9671E-04 & 1.3546E-05 &       & 4.3184E-03 & 3.9726E-06 &       & 5.6020E-08 & 3.1823E-08 &  \\
				\bottomrule
			\end{tabular}%
			
		}%
	\end{minipage}
\end{table*}

\begin{table*}[htbp]
	\renewcommand{\arraystretch}{0.5}
	\caption{DMOP2$_{iso}$}
	\label{Results:DMOP2_iso}
	\centering
	\begin{minipage}[t]{0.85\textwidth}%
		\resizebox{0.99\textwidth}{!}{
			\begin{tabular}{c|cccccccccc}
				\toprule
				\toprule
				&       & { {\scriptsize NSGA-II} }   & {\scriptsize Tr-NSGA-II} & \scriptsize{ROC (\%)}   & {\scriptsize MOPSO }    & {\scriptsize Tr-MOPSO } & \scriptsize{ROC (\%)}  & {\scriptsize RM-MEDA }    & Tr-RM-MEDA  & \scriptsize{ROC (\%)} \\
				\midrule
				\multirow{2}[2]{*}{\textbf{C1}} & MIGD  & 0.0026 & 0.0027 & \multirow{2}[2]{*}{-4.8403 } & 0.0048 & 0.0049 & \multirow{2}[2]{*}{-0.2598 } & 0.0019 & 0.0019 & \multirow{2}[2]{*}{-0.0057} \\
				& VAR   & 1.4710E-12 & 7.0791E-10 &       & 9.0873E-09 & 3.2098E-10 &       & 5.4961E-14 & 1.0336E-11 &  \\
				\midrule
				\multirow{2}[2]{*}{\textbf{C2}} & MIGD  & 0.0026 & 0.0027 & \multirow{2}[2]{*}{-4.4605 } & 0.0050 & 0.0049 & \multirow{2}[2]{*}{\textbf{2.5526 }} & 0.0019 & 0.0019 & \multirow{2}[2]{*}{-0.0855} \\
				& VAR   & 7.0355E-10 & 2.1892E-10 &       & 8.0595E-09 & 6.0198E-11 &       & 4.1426E-12 & 3.8878E-13 &  \\
				\midrule
				\multirow{2}[2]{*}{\textbf{C3}} & MIGD  & 0.0026 & 0.0027 & \multirow{2}[2]{*}{-6.8106 } & 0.0049 & 0.0048 & \multirow{2}[2]{*}{\textbf{1.6411 }} & 0.0019 & 0.0019 & \multirow{2}[2]{*}{-0.0549} \\
				& VAR   & 1.4253E-09 & 2.3627E-09 &       & 9.2590E-09 & 1.0770E-08 &       & 5.3240E-13 & 3.5049E-12 &  \\
				\midrule
				\multirow{2}[2]{*}{\textbf{C4}} & MIGD  & 0.0026 & 0.0027 & \multirow{2}[2]{*}{-3.1781 } & 0.0051 & 0.0050 & \multirow{2}[2]{*}{\textbf{2.8216 }} & 0.0019 & 0.0019 & \multirow{2}[2]{*}{-0.0609} \\
				& VAR   & 4.3446E-11 & 5.6923E-09 &       & 8.3902E-13 & 3.4667E-09 &       & 2.7556E-12 & 1.2409E-13 &  \\
				\midrule
				\multirow{2}[2]{*}{\textbf{C5}} & MIGD  & 0.1212 & 0.1304 & \multirow{2}[2]{*}{-7.5446 } & 0.1127 & 0.1127 & \multirow{2}[2]{*}{-0.0139 } & 0.1104 & 0.1104 & \multirow{2}[2]{*}{\textbf{0.0017}} \\
				& VAR   & 1.4153E-05 & 1.2108E-05 &       & 5.5578E-08 & 6.6161E-10 &       & 3.0130E-12 & 7.3002E-12 &  \\
				\midrule
				\multirow{2}[2]{*}{\textbf{C6}} & MIGD  & 0.1232 & 0.1402 & \multirow{2}[2]{*}{-13.7891 } & 0.1127 & 0.1128 & \multirow{2}[2]{*}{-0.0400 } & 0.1104 & 0.1104 & \multirow{2}[2]{*}{-0.0004} \\
				& VAR   & 1.4552E-05 & 2.1830E-06 &       & 4.6055E-08 & 1.7908E-08 &       & 7.9401E-16 & 2.4929E-12 &  \\
				\midrule
				\multirow{2}[2]{*}{\textbf{C7}} & MIGD  & 0.0025 & 0.0026 & \multirow{2}[2]{*}{-2.9765 } & 0.0048 & 0.0049 & \multirow{2}[2]{*}{-0.5804 } & 0.0019 & 0.0019 & \multirow{2}[2]{*}{\textbf{0.0712}} \\
				& VAR   & 5.2352E-10 & 3.4636E-10 &       & 1.3988E-08 & 2.4884E-09 &       & 4.2732E-12 & 1.5004E-12 &  \\
				\midrule
				\multirow{2}[2]{*}{\textbf{C8}} & MIGD  & 0.0026 & 0.0026 & \multirow{2}[2]{*}{-2.4833 } & 0.0050 & 0.0049 & \multirow{2}[2]{*}{\textbf{3.3136 }} & 0.0019 & 0.0019 & \multirow{2}[2]{*}{-0.0935} \\
				& VAR   & 1.1559E-10 & 8.7427E-10 &       & 1.6665E-08 & 3.8434E-08 &       & 6.8648E-12 & 2.1063E-12 &  \\
				\bottomrule	
			\end{tabular}				
			
		}%
	\end{minipage}
	
	\renewcommand{\arraystretch}{0.5}
	\caption{DMOP2$_{dec}$}
	\label{Results:DMOP2_dec}
	\centering
	\begin{minipage}[t]{0.85\textwidth}%
		\resizebox{0.99\textwidth}{!}{
			\begin{tabular}{c|cccccccccc}
				\toprule
				\toprule
				&       & { {\scriptsize NSGA-II} }   & {\scriptsize Tr-NSGA-II} & \scriptsize{ROC (\%)}   & {\scriptsize MOPSO }    & {\scriptsize Tr-MOPSO } & \scriptsize{ROC (\%)}  & {\scriptsize RM-MEDA }    & Tr-RM-MEDA  & \scriptsize{ROC (\%)} \\
				\midrule
				\multirow{2}[2]{*}{\textbf{C1}} & MIGD  & 0.3575 & 0.1969 & \multirow{2}[2]{*}{\textbf{44.9115 }} & 0.3054 & 0.0307 & \multirow{2}[2]{*}{\textbf{89.9592 }} & 0.1104 & 0.0511 & \multirow{2}[2]{*}{\textbf{53.7191}} \\
				& VAR   & 5.4036E-03 & 8.4656E-04 &       & 5.7504E-02 & 4.0371E-05 &       & 4.5859E-05 & 4.4133E-06 &  \\
				\midrule
				\multirow{2}[2]{*}{\textbf{C2}} & MIGD  & 0.4193 & 0.2044 & \multirow{2}[2]{*}{\textbf{51.2453 }} & 0.2230 & 0.0134 & \multirow{2}[2]{*}{\textbf{93.9837 }} & 0.1465 & 0.0523 & \multirow{2}[2]{*}{\textbf{64.3164}} \\
				& VAR   & 3.3718E-03 & 1.3931E-06 &       & 7.8028E-05 & 1.0865E-06 &       & 2.0992E-04 & 4.1956E-05 &  \\
				\midrule
				\multirow{2}[2]{*}{\textbf{C3}} & MIGD  & 0.3923 & 0.2182 & \multirow{2}[2]{*}{\textbf{44.3910 }} & 0.2267 & 0.0138 & \multirow{2}[2]{*}{\textbf{93.9166 }} & 0.0912 & 0.0446 & \multirow{2}[2]{*}{\textbf{51.0528}} \\
				& VAR   & 2.2371E-03 & 3.4360E-06 &       & 3.7621E-03 & 4.9494E-08 &       & 2.5922E-05 & 8.8813E-07 &  \\
				\midrule
				\multirow{2}[2]{*}{\textbf{C4}} & MIGD  & 0.4062 & 0.2432 & \multirow{2}[2]{*}{\textbf{40.1262 }} & 0.4104 & 0.0142 & \multirow{2}[2]{*}{\textbf{96.5336 }} & 0.1104 & 0.0441 & \multirow{2}[2]{*}{\textbf{60.0153}} \\
				& VAR   & 1.5897E-03 & 1.2597E-03 &       & 6.8751E-02 & 2.4467E-06 &       & 1.6155E-04 & 4.1179E-05 &  \\
				\midrule
				\multirow{2}[2]{*}{\textbf{C5}} & MIGD  & 1.2991 & 0.9970 & \multirow{2}[2]{*}{\textbf{23.2553 }} & 0.6897 & 0.0514 & \multirow{2}[2]{*}{\textbf{92.5402 }} & 0.2322 & 0.2290 & \multirow{2}[2]{*}{\textbf{1.3929}} \\
				& VAR   & 5.5417E-02 & 7.5395E-06 &       & 1.4321E-03 & 1.1253E-05 &       & 1.6904E-05 & 1.2066E-04 &  \\
				\midrule
				\multirow{2}[2]{*}{\textbf{C6}} & MIGD  & 1.5098 & 0.8689 & \multirow{2}[2]{*}{\textbf{42.4497 }} & 0.6992 & 0.0553 & \multirow{2}[2]{*}{\textbf{92.0972 }} & 0.2301 & 0.2213 & \multirow{2}[2]{*}{\textbf{3.8511}} \\
				& VAR   & 2.8521E-02 & 1.6001E-03 &       & 2.3515E-01 & 2.8605E-05 &       & 3.4389E-05 & 7.2266E-06 &  \\
				\midrule
				\multirow{2}[2]{*}{\textbf{C7}} & MIGD  & 0.3701 & 0.2011 & \multirow{2}[2]{*}{\textbf{45.6631 }} & 0.4014 & 0.0115 & \multirow{2}[2]{*}{\textbf{97.1391 }} & 0.1315 & 0.0553 & \multirow{2}[2]{*}{\textbf{57.9290}} \\
				& VAR   & 3.8975E-03 & 4.6698E-03 &       & 2.3974E-05 & 9.5240E-08 &       & 5.0617E-05 & 4.0637E-05 &  \\
				\midrule
				\multirow{2}[2]{*}{\textbf{C8}} & MIGD  & 0.2878 & 0.2142 & \multirow{2}[2]{*}{\textbf{25.5755 }} & 0.3977 & 0.0133 & \multirow{2}[2]{*}{\textbf{96.6628 }} & 0.1066 & 0.0540 & \multirow{2}[2]{*}{\textbf{49.2849}} \\
				& VAR   & 1.7600E-03 & 1.8550E-04 &       & 2.0118E-02 & 7.5405E-07 &       & 5.8950E-05 & 8.2834E-05 &  \\
				\bottomrule
			\end{tabular}%
			
		}%
	\end{minipage}
\end{table*}

\begin{table*}[htbp]
	\renewcommand{\arraystretch}{0.5}
	\caption{DMOP3}
	\label{Results:DMOP3}
	\centering
	\begin{minipage}[t]{0.85\textwidth}%
		\resizebox{0.99\textwidth}{!}{
			\begin{tabular}{c|cccccccccc}
				\toprule
				\toprule
				&       & { {\scriptsize NSGA-II} }   & {\scriptsize Tr-NSGA-II} & \scriptsize{ROC (\%)}   & {\scriptsize MOPSO }    & {\scriptsize Tr-MOPSO } & \scriptsize{ROC (\%)}  & {\scriptsize RM-MEDA }    & Tr-RM-MEDA  & \scriptsize{ROC (\%)} \\
				\midrule
				\multirow{2}[2]{*}{\textbf{C1}} & MIGD  & 0.1084 & 0.0296 & \multirow{2}[2]{*}{\textbf{72.7002 }} & 0.0105 & 0.0416 & \multirow{2}[2]{*}{-296.0364 } & 0.0033 & \multirow{2}[2]{*}{0.0025} & \multirow{2}[2]{*}{\textbf{24.6490}} \\
				& VAR   & 7.3738E-05 & 7.9556E-06 &       & 1.6573E-05 & 2.4358E-03 &       & 4.5813E-11 &       &  \\
				\midrule
				\multirow{2}[2]{*}{\textbf{C2}} & MIGD  & 0.1359 & 0.0359 & \multirow{2}[2]{*}{\textbf{73.6077 }} & 0.0143 & 0.0065 & \multirow{2}[2]{*}{\textbf{54.2790 }} & 0.0031 & \multirow{2}[2]{*}{0.0024} & \multirow{2}[2]{*}{\textbf{22.0849}} \\
				& VAR   & 2.3903E-04 & 2.2027E-04 &       & 4.9260E-05 & 5.5000E-09 &       & 6.8391E-09 &       &  \\
				\midrule
				\multirow{2}[2]{*}{\textbf{C3}} & MIGD  & 0.1486 & 0.0245 & \multirow{2}[2]{*}{\textbf{83.5041 }} & 0.0077 & 0.0062 & \multirow{2}[2]{*}{\textbf{19.3390 }} & 0.0030 & \multirow{2}[2]{*}{0.0038} & \multirow{2}[2]{*}{-25.1478} \\
				& VAR   & 1.0447E-06 & 3.2505E-05 &       & 3.1252E-07 & 8.9046E-09 &       & 3.3853E-09 &       &  \\
				\midrule
				\multirow{2}[2]{*}{\textbf{C4}} & MIGD  & 0.0922 & 0.0390 & \multirow{2}[2]{*}{\textbf{57.6545 }} & 0.0090 & 0.0064 & \multirow{2}[2]{*}{\textbf{28.1404 }} & 0.0030 & \multirow{2}[2]{*}{0.0030} & \multirow{2}[2]{*}{\textbf{0.4896}} \\
				& VAR   & 7.5602E-05 & 2.6008E-08 &       & 7.6625E-09 & 1.0967E-08 &       & 1.0279E-08 &       &  \\
				\midrule
				\multirow{2}[2]{*}{\textbf{C5}} & MIGD  & 0.3562 & 2.1887 & \multirow{2}[2]{*}{-514.5040 } & 0.1310 & 1.9281 & \multirow{2}[2]{*}{-1371.9864 } & 18.3638 & \multirow{2}[2]{*}{18.4456} & \multirow{2}[2]{*}{-0.4455} \\
				& VAR   & 1.0547E-03 & 7.7773E+00 &       & 4.5042E-04 & 6.4553E+00 &       & 3.4346E-03 &       &  \\
				\midrule
				\multirow{2}[2]{*}{\textbf{C6}} & MIGD  & 5.9830 & 5.7247 & \multirow{2}[2]{*}{\textbf{4.3171 }} & 2.0842 & 0.1169 & \multirow{2}[2]{*}{\textbf{94.3914 }} & 18.3348 & \multirow{2}[2]{*}{18.4792} & \multirow{2}[2]{*}{-0.7879} \\
				& VAR   & 8.8697E+00 & 8.1335E+00 &       & 7.5217E+00 & 2.9454E-06 &       & 8.5653E-03 &       &  \\
				\midrule
				\multirow{2}[2]{*}{\textbf{C7}} & MIGD  & 0.1395 & 0.0360 & \multirow{2}[2]{*}{\textbf{74.1808 }} & 0.0155 & 0.0076 & \multirow{2}[2]{*}{\textbf{51.0026 }} & 0.0029 & \multirow{2}[2]{*}{0.0026} & \multirow{2}[2]{*}{\textbf{10.7489}} \\
				& VAR   & 3.9562E-04 & 5.3965E-06 &       & 1.0151E-04 & 2.2629E-08 &       & 4.1482E-09 &       &  \\
				\midrule
				\multirow{2}[2]{*}{\textbf{C8}} & MIGD  & 0.1174 & 0.0278 & \multirow{2}[2]{*}{\textbf{76.2864 }} & 0.0086 & 0.0063 & \multirow{2}[2]{*}{\textbf{26.7219 }} & 0.0033 & \multirow{2}[2]{*}{0.0024} & \multirow{2}[2]{*}{\textbf{27.4789}} \\
				& VAR   & 1.1919E-04 & 1.0728E-04 &       & 1.9468E-06 & 2.4955E-07 &       & 3.5712E-07 &       &  \\
				\bottomrule
			\end{tabular}%
			
		}%
	\end{minipage}
	
	\renewcommand{\arraystretch}{0.5}
	\caption{HE2}
	\label{Results:HE2}
	\centering
	\begin{minipage}[t]{0.85\textwidth}%
		\resizebox{0.99\textwidth}{!}{
			\begin{tabular}{c|cccccccccc}
				\toprule
				\toprule
				&       & { {\scriptsize NSGA-II} }   & {\scriptsize Tr-NSGA-II} & \scriptsize{ROC (\%)}   & {\scriptsize MOPSO }    & {\scriptsize Tr-MOPSO } & \scriptsize{ROC (\%)}  & {\scriptsize RM-MEDA }    & Tr-RM-MEDA  & \scriptsize{ROC (\%)} \\
				\midrule
				\multirow{2}[2]{*}{\textbf{C1}} & MIGD  & 0.1942 & 0.1889 & \multirow{2}[2]{*}{\textbf{2.7314 }} & 0.0795 & 0.0683 & \multirow{2}[2]{*}{\textbf{14.0445 }} & 0.8809 & \multirow{2}[2]{*}{0.1162} & \multirow{2}[2]{*}{\textbf{86.8097}} \\
				& VAR   & 1.2666E-03 & 1.7182E-04 &       & 1.9892E-04 & 1.6164E-05 &       & 9.0862E-05 &       &  \\
				\midrule
				\multirow{2}[2]{*}{\textbf{C2}} & MIGD  & 0.1522 & 0.1788 & \multirow{2}[2]{*}{-17.4533 } & 0.1019 & 0.0672 & \multirow{2}[2]{*}{\textbf{34.0322 }} & 0.9015 & \multirow{2}[2]{*}{0.1171} & \multirow{2}[2]{*}{\textbf{87.0121}} \\
				& VAR   & 1.3660E-04 & 6.1378E-04 &       & 2.6750E-03 & 1.4320E-05 &       & 3.6855E-04 &       &  \\
				\midrule
				\multirow{2}[2]{*}{\textbf{C3}} & MIGD  & 0.1572 & 0.1546 & \multirow{2}[2]{*}{\textbf{1.6659 }} & 0.0784 & 0.0697 & \multirow{2}[2]{*}{\textbf{11.1071 }} & 0.9025 & \multirow{2}[2]{*}{0.1196} & \multirow{2}[2]{*}{\textbf{86.7433}} \\
				& VAR   & 1.6637E-06 & 4.0007E-05 &       & 3.2323E-04 & 1.1259E-05 &       & 1.0819E-04 &       &  \\
				\midrule
				\multirow{2}[2]{*}{\textbf{C4}} & MIGD  & 0.1875 & 0.1765 & \multirow{2}[2]{*}{\textbf{5.8589 }} & 0.0949 & 0.0623 & \multirow{2}[2]{*}{\textbf{34.3376 }} & 0.8906 & \multirow{2}[2]{*}{0.1068} & \multirow{2}[2]{*}{\textbf{88.0113}} \\
				& VAR   & 5.7780E-04 & 4.0523E-05 &       & 1.7837E-03 & 4.6907E-07 &       & 1.1358E-04 &       &  \\
				\midrule
				\multirow{2}[2]{*}{\textbf{C5}} & MIGD  & 0.2700 & 0.0879 & \multirow{2}[2]{*}{\textbf{67.4516 }} & 0.0967 & 0.0595 & \multirow{2}[2]{*}{\textbf{38.4635 }} & 0.6971 & \multirow{2}[2]{*}{0.0697} & \multirow{2}[2]{*}{\textbf{89.9980}} \\
				& VAR   & 9.1661E-04 & 1.4581E-04 &       & 2.5397E-04 & 1.3812E-05 &       & 1.1198E-06 &       &  \\
				\midrule
				\multirow{2}[2]{*}{\textbf{C6}} & MIGD  & 0.3545 & 0.0815 & \multirow{2}[2]{*}{\textbf{77.0146 }} & 0.0894 & 0.0572 & \multirow{2}[2]{*}{\textbf{36.0087 }} & 0.6974 & \multirow{2}[2]{*}{0.0726} & \multirow{2}[2]{*}{\textbf{89.5874}} \\
				& VAR   & 1.2909E-03 & 1.5771E-04 &       & 2.3229E-04 & 9.3198E-08 &       & 2.5975E-05 &       &  \\
				\midrule
				\multirow{2}[2]{*}{\textbf{C7}} & MIGD  & 0.2078 & 0.1495 & \multirow{2}[2]{*}{\textbf{28.0723 }} & 0.0638 & 0.0650 & \multirow{2}[2]{*}{-1.9613 } & 0.8962 & \multirow{2}[2]{*}{0.1038} & \multirow{2}[2]{*}{\textbf{88.4222}} \\
				& VAR   & 4.0462E-06 & 1.0892E-04 &       & 9.4058E-07 & 6.6609E-06 &       & 7.6239E-07 &       &  \\
				\midrule
				\multirow{2}[2]{*}{\textbf{C8}} & MIGD  & 0.1531 & 0.1830 & \multirow{2}[2]{*}{-19.5666 } & 0.0728 & 0.0626 & \multirow{2}[2]{*}{\textbf{14.0586 }} & 0.8944 & \multirow{2}[2]{*}{0.1082} & \multirow{2}[2]{*}{\textbf{87.9062}} \\
				& VAR   & 2.3195E-04 & 5.1881E-04 &       & 4.2019E-05 & 2.7083E-07 &       & 3.6057E-04 &       &  \\
				\bottomrule
			\end{tabular}%
			
		}%
	\end{minipage}
\end{table*}

\begin{table*}[htbp]
	\renewcommand{\arraystretch}{0.5}
	\caption{HE7}
	\label{Results:HE7}
	\centering
	\begin{minipage}[t]{0.85\textwidth}%
		\resizebox{0.99\textwidth}{!}{
			\begin{tabular}{c|cccccccccc}
				\toprule
				\toprule
				&       & { {\scriptsize NSGA-II} }   & {\scriptsize Tr-NSGA-II} & \scriptsize{ROC (\%)}   & {\scriptsize MOPSO }    & {\scriptsize Tr-MOPSO } & \scriptsize{ROC (\%)}  & {\scriptsize RM-MEDA }    & Tr-RM-MEDA  & \scriptsize{ROC (\%)} \\
				\midrule
				\multirow{2}[2]{*}{\textbf{C1}} & MIGD  & 0.0971 & 0.0417 & \multirow{2}[2]{*}{\textbf{57.1039 }} & 0.0612 & 0.0594 & \multirow{2}[2]{*}{\textbf{2.8756 }} & 0.0438 & 0.0354 & \multirow{2}[2]{*}{\textbf{19.3090}} \\
				& VAR   & 1.4370E-06 & 1.7339E-06 &       & 2.9389E-04 & 3.4081E-06 &       & 1.7793E-04 & 4.5205E-06 &  \\
				\midrule
				\multirow{2}[2]{*}{\textbf{C2}} & MIGD  & 0.0988 & 0.0409 & \multirow{2}[2]{*}{\textbf{58.5736 }} & 0.0588 & 0.0608 & \multirow{2}[2]{*}{-3.3986 } & 0.0440 & 0.0344 & \multirow{2}[2]{*}{\textbf{21.8204}} \\
				& VAR   & 1.8807E-06 & 2.0477E-06 &       & 7.8924E-05 & 6.3404E-08 &       & 1.7464E-04 & 1.9297E-06 &  \\
				\midrule
				\multirow{2}[2]{*}{\textbf{C3}} & MIGD  & 0.1012 & 0.0410 & \multirow{2}[2]{*}{\textbf{59.4455 }} & 0.0600 & 0.0611 & \multirow{2}[2]{*}{-1.7770 } & 0.0449 & 0.0353 & \multirow{2}[2]{*}{\textbf{21.4040}} \\
				& VAR   & 1.0019E-05 & 2.4940E-10 &       & 5.2104E-05 & 1.3924E-06 &       & 1.5950E-04 & 1.9108E-07 &  \\
				\midrule
				\multirow{2}[2]{*}{\textbf{C4}} & MIGD  & 0.1036 & 0.0398 & \multirow{2}[2]{*}{\textbf{61.5666 }} & 0.0596 & 0.0598 & \multirow{2}[2]{*}{-0.4940 } & 0.0440 & 0.0351 & \multirow{2}[2]{*}{\textbf{20.2879}} \\
				& VAR   & 3.2259E-05 & 1.9226E-06 &       & 3.2056E-05 & 3.3301E-06 &       & 1.7977E-04 & 1.4912E-06 &  \\
				\midrule
				\multirow{2}[2]{*}{\textbf{C5}} & MIGD  & 0.0779 & 0.0362 & \multirow{2}[2]{*}{\textbf{53.4686 }} & 0.0613 & 0.0528 & \multirow{2}[2]{*}{\textbf{13.9572 }} & 0.0397 & 0.0321 & \multirow{2}[2]{*}{\textbf{19.2266}} \\
				& VAR   & 2.4013E-05 & 9.3120E-08 &       & 7.4316E-05 & 1.4570E-05 &       & 1.1893E-04 & 4.0164E-06 &  \\
				\midrule
				\multirow{2}[2]{*}{\textbf{C6}} & MIGD  & 0.0781 & 0.0326 & \multirow{2}[2]{*}{\textbf{58.3400 }} & 0.0579 & 0.0532 & \multirow{2}[2]{*}{\textbf{8.1117 }} & 0.0391 & 0.0333 & \multirow{2}[2]{*}{\textbf{14.8676}} \\
				& VAR   & 7.6260E-06 & 6.2165E-08 &       & 1.5637E-05 & 2.0849E-05 &       & 1.4266E-04 & 6.3904E-06 &  \\
				\midrule
				\multirow{2}[2]{*}{\textbf{C7}} & MIGD  & 0.0939 & 0.0396 & \multirow{2}[2]{*}{\textbf{57.7739 }} & 0.0523 & 0.0599 & \multirow{2}[2]{*}{-14.4283 } & 0.0433 & 0.0340 & \multirow{2}[2]{*}{\textbf{21.3226}} \\
				& VAR   & 3.4045E-05 & 1.9975E-06 &       & 1.2234E-05 & 1.8850E-06 &       & 1.3745E-04 & 4.2448E-07 &  \\
				\midrule
				\multirow{2}[2]{*}{\textbf{C8}} & MIGD  & 0.1059 & 0.0403 & \multirow{2}[2]{*}{\textbf{61.9792 }} & 0.0547 & 0.0597 & \multirow{2}[2]{*}{-9.1221 } & 0.0436 & 0.0337 & \multirow{2}[2]{*}{\textbf{22.5674}} \\
				& VAR   & 2.5446E-06 & 4.2571E-06 &       & 2.3103E-06 & 2.1898E-06 &       & 1.5031E-04 & 4.0268E-06 &  \\
				\bottomrule
			\end{tabular}%
			
		}%
	\end{minipage}
	
	\renewcommand{\arraystretch}{0.5}
	\caption{HE9}
	\label{Results:HE9}
	\centering
	\begin{minipage}[t]{0.85\textwidth}%
		\resizebox{0.99\textwidth}{!}{
			\begin{tabular}{c|cccccccccc}
				\toprule
				\toprule
				&       & { {\scriptsize NSGA-II} }   & {\scriptsize Tr-NSGA-II} & \scriptsize{ROC (\%)}   & {\scriptsize MOPSO }    & {\scriptsize Tr-MOPSO } & \scriptsize{ROC (\%)}  & {\scriptsize RM-MEDA }    & Tr-RM-MEDA  & \scriptsize{ROC (\%)} \\
				\midrule
				\multirow{2}[2]{*}{\textbf{C1}} & MIGD  & 0.3028 & 0.2580 & \multirow{2}[2]{*}{\textbf{14.7877 }} & 0.2648 & 0.2961 & \multirow{2}[2]{*}{-11.8097 } & 0.2653 & 0.2455 & \multirow{2}[2]{*}{\textbf{7.4804 }} \\
				& VAR   & 2.9107E-05 & 1.5525E-06 &       & 8.0165E-06 & 3.5768E-05 &       & 4.0274E-08 & 3.9133E-07 &  \\
				\midrule
				\multirow{2}[2]{*}{\textbf{C2}} & MIGD  & 0.3009 & 0.2617 & \multirow{2}[2]{*}{\textbf{13.0234 }} & 0.2712 & 0.3023 & \multirow{2}[2]{*}{-11.4461 } & 0.2637 & 0.2437 & \multirow{2}[2]{*}{\textbf{7.6087 }} \\
				& VAR   & 2.0520E-06 & 5.9900E-06 &       & 7.1836E-07 & 4.6393E-05 &       & 3.1215E-07 & 2.4694E-07 &  \\
				\midrule
				\multirow{2}[2]{*}{\textbf{C3}} & MIGD  & 0.3057 & 0.2548 & \multirow{2}[2]{*}{\textbf{16.6490 }} & 0.2675 & 0.3004 & \multirow{2}[2]{*}{-12.2867 } & 0.2634 & 0.2451 & \multirow{2}[2]{*}{\textbf{6.9490 }} \\
				& VAR   & 2.2473E-05 & 2.2159E-05 &       & 1.6559E-06 & 2.1955E-05 &       & 1.4103E-06 & 3.2609E-07 &  \\
				\midrule
				\multirow{2}[2]{*}{\textbf{C4}} & MIGD  & 0.3029 & 0.2635 & \multirow{2}[2]{*}{\textbf{12.9993 }} & 0.2655 & 0.3028 & \multirow{2}[2]{*}{-14.0407 } & 0.2639 & 0.2457 & \multirow{2}[2]{*}{\textbf{6.9055 }} \\
				& VAR   & 4.6335E-05 & 8.2176E-05 &       & 1.0313E-06 & 2.9225E-05 &       & 3.7052E-06 & 2.2156E-05 &  \\
				\midrule
				\multirow{2}[2]{*}{\textbf{C5}} & MIGD  & 0.2670 & 0.2270 & \multirow{2}[2]{*}{\textbf{14.9754 }} & 0.1877 & 0.2572 & \multirow{2}[2]{*}{-37.0047 } & 0.2365 & 0.2168 & \multirow{2}[2]{*}{\textbf{8.3525 }} \\
				& VAR   & 1.1548E-06 & 1.5988E-06 &       & 5.1013E-06 & 1.0775E-04 &       & 1.3658E-05 & 5.3162E-10 &  \\
				\midrule
				\multirow{2}[2]{*}{\textbf{C6}} & MIGD  & 0.2669 & 0.2279 & \multirow{2}[2]{*}{\textbf{14.6107 }} & 0.1908 & 0.2544 & \multirow{2}[2]{*}{-33.2995 } & 0.2356 & 0.2179 & \multirow{2}[2]{*}{\textbf{7.5252 }} \\
				& VAR   & 3.1621E-06 & 2.7471E-05 &       & 4.2148E-08 & 2.3717E-05 &       & 2.7905E-07 & 5.3859E-06 &  \\
				\midrule
				\multirow{2}[2]{*}{\textbf{C7}} & MIGD  & 0.3050 & 0.2558 & \multirow{2}[2]{*}{\textbf{16.1353 }} & 0.2551 & 0.2963 & \multirow{2}[2]{*}{-16.1533 } & 0.2626 & 0.2449 & \multirow{2}[2]{*}{\textbf{6.7409 }} \\
				& VAR   & 8.0651E-05 & 1.5958E-05 &       & 3.2539E-07 & 4.7023E-05 &       & 3.7952E-06 & 1.0545E-05 &  \\
				\midrule
				\multirow{2}[2]{*}{\textbf{C8}} & MIGD  & 0.3051 & 0.2579 & \multirow{2}[2]{*}{\textbf{15.4894 }} & 0.2645 & 0.3003 & \multirow{2}[2]{*}{-13.5733 } & 0.2609 & 0.2467 & \multirow{2}[2]{*}{\textbf{5.4267 }} \\
				& VAR   & 2.4933E-05 & 3.7407E-06 &       & 1.2709E-07 & 1.0673E-05 &       & 5.5975E-07 & 5.4257E-09 &  \\
				\bottomrule
			\end{tabular}%
			
		}%
	\end{minipage}
\end{table*}
\end{document}